\documentclass[10pt,twocolumn,letterpaper]{article}

\usepackage{cvpr}
\usepackage{times}
\usepackage{epsfig}
\usepackage{graphicx}
\usepackage{amsmath}
\usepackage{amssymb}
\usepackage{amsthm}
\usepackage{multirow}
\usepackage{url}
\usepackage{xcolor}
 
\usepackage{caption}
\usepackage{subcaption}

\newtheorem{theorem}{Theorem}

\usepackage{alphalph}

\usepackage{appendix}

\usepackage{amsmath,amsfonts,bm}

\def\eqref#1{equation~\ref{#1}}
\def\1{\bm{1}}

\def\vn{{\bm{n}}}

\DeclareMathAlphabet{\mathsfit}{\encodingdefault}{\sfdefault}{m}{sl}
\SetMathAlphabet{\mathsfit}{bold}{\encodingdefault}{\sfdefault}{bx}{n}

\DeclareMathOperator*{\argmin}{arg\,min}

\usepackage[pagebackref=true,breaklinks=true,letterpaper=true,colorlinks,bookmarks=false]{hyperref}

\cvprfinalcopy % *** Uncomment this line for the final submission

\def\cvprPaperID{7522} % *** Enter the CVPR Paper ID here

\ifcvprfinal\fi
\begin{document}

%%%%%%%%% TITLE
\title{Visualizing Point Cloud Classifiers by Curvature Smoothing}

\author{Chen Ziwen\\
Grinnell College\\
Grinnell, IA\\
{\tt\small chenziwe@grinnell.edu}
\and
Wenxuan Wu\\
Oregon State University\\
Corvallis, OR\\
{\tt\small wuwen@oregonstate.edu}
\and
Zhongang Qi\\
Applied Research Center (ARC)\\
Tencent PCG\\
Shenzhen, China\\
{\tt\small zhongangqi@tencent.com}
\and
Li Fuxin\\
Oregon State University\\
Corvallis, OR\\
{\tt\small lif@oregonstate.edu}
}

\maketitle

\begin{abstract}
Recently, several networks that operate directly on point clouds have been proposed. There is significant utility in understanding their mechanisms to classify point clouds, which can potentially help diagnosing these networks and designing better architectures.  In this paper, we propose a novel approach to visualize features important to the point cloud classifiers.
Our approach is based on smoothing curved areas on a point cloud. After prominent features were smoothed, the resulting point cloud can be evaluated on the network to assess whether the feature is important to the classifier. A technical contribution of the paper is an approximated curvature smoothing algorithm, which can smoothly transition from the original point cloud to one of constant curvature, such as a uniform sphere. Based on the smoothing algorithm, we propose PCI-GOS (Point Cloud Integrated-Gradients Optimized Saliency), a visualization technique that can automatically find the minimal saliency map that covers the most important features on a shape. Experiment results revealed insights into different point cloud classifiers. The code is available at \url{https://github.com/arthurhero/PC-IGOS} \footnote{This work was done while Zhongang Qi was a Postdoctoral Scholar at Oregon State University}
\end{abstract}

%-------------------------------------------------------------------------

\begin{figure*}[htb]
\begin{subfigure}{.48\linewidth}
\includegraphics[width=0.95\linewidth]{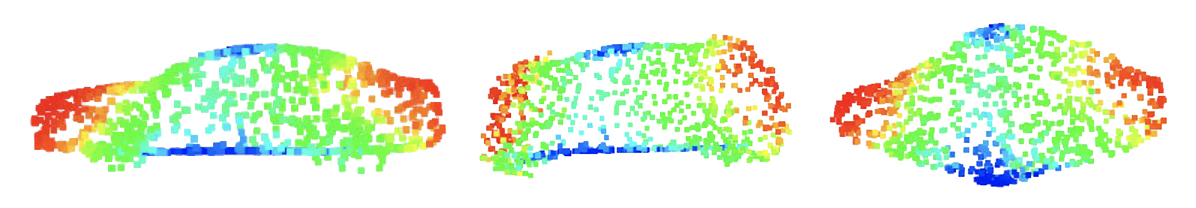}
\captionof{figure}{Car: 1.00, 0.09, 0.99.}
\end{subfigure}\hfill
\begin{subfigure}{.48\linewidth}
\includegraphics[width=0.95\linewidth]{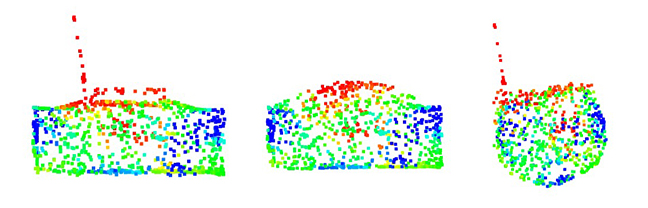}
\captionof{figure}{Radio: 0.77, 0.02, 0.92}
\end{subfigure}
\vspace{0.1in}
\caption{We visualize point cloud classifiers by smoothing curved areas. The numbers show the prediction confidence of each cloud. From left to right: Original point cloud, minimally smoothed for a predicted confidence of less than 10\% of the original, maximally smoothed for a predicted confidence of more than 90\% of the original.}
\vskip -0.15in
\label{fig1}
\end{figure*}

\section{Introduction}
\label{sec:intro}
Recently, direct deep learning on unstructured 3-D point clouds has gained significant interest. Many interesting point cloud networks have been proposed. PointNet++ \cite{pointnet2} utilizes max-pooling followed by multi-layer perceptron.
PointConv \cite{pointconv} realizes a convolution operation on point clouds efficiently. Other works such as \cite{dgcnn, splatnet,spider,pcnn,pointcnn,splinecnn,tangentnet} all have their own merits. As with 2-D image classifiers, we are curious about what  these models have actually learned. 
%In other words, we want to \textit{explain} those models by identifying which parts of a shape contribute most to the final score, and which parts the least. 
Such explanations would help us gain more insights, diagnose the networks, and potentially design better network structures and data augmentation pipelines.

In this work, we are interested in looking for the most important features on a shape for the classifiers. Following the \textit{deletion} and \textit{insertion} metric proposed by \cite{deletion}, we should expect  the predicted score to drop quickly when we ``cover up" those important features, and to rise quickly when we gradually ``reveal" \textit{only} those important features. We want to design an algorithm that can automatically learn the minimal saliency map as in \cite{mask}.

In order to apply a saliency map on a shape, we need an operator that can \textit{gradually} ``cover up" and ``reveal" parts of a point cloud. For 2-D images we can simply apply different levels of Gaussian blur to the pixels. However in 3-D, no matter how we move the points, they will always be part of the point cloud, and thus contributing to the underlying shape. %We can directly delete points as in prior work \cite{saliency}, but the classifier might not be robust to direct deletion.
%Our goal is to perform this ``cover up" process in a manner so that the resulting point cloud has the same number of points as before, and does not create irregular shapes that fall outside the training distribution. Hence, we attempt to smoothly morph the 3-D shape to remove distinctive shape features, such as edges and corners. With the key observation that edges and corners are reflected by abnormal curvatures on the underlying surface, we specify the goal of our algorithm to be averaging the curvature on the underlying surface of the point cloud (thus the final shape should have approximately constant curvature everywhere).
With the key observation that sharp features like edges and corners  on a shape are reflected by abnormal curvatures on the underlying surface, we propose a novel, diffusion-based smoothing algorithm that can gradually smooth out curvatures on a point cloud. 
%For each point in the point cloud, a local plane is fit to its neighborhood. We prove under some conditions that the distance from the point to this local plane is a valid approximation of the local mean curvature (see supplementary material). This allows us to utilize a  smoothing algorithm, which 
%Our algorithm can gradually average out the approximated curvature on the underlying surface. 
For instance, if the underlying surface is closed, then our algorithm will gradually morph the shape into a sphere.

With the smoothing method, we propose PCI-GOS (``point-cloud I-GOS"), a 3-D heatmap visualization algorithm. This extends the I-GOS algorithm~\cite{igos} on 2D images to generate a saliency map that highlights points which are important for classifiers.
We experiment our approach on PointConv \cite{pointconv}, a state-of-the-art point cloud network. We compare our results on the ModelNet40 dataset with several baselines including Zheng \textit{et al.} \cite{saliency}, a gradient-based visualization technique optimized for direct point deletion.%, with metrics proposed by \cite{deletion} and with qualitative illustrations.
%reveals that, different from image-based networks that often classify based on a small distinctive feature, point-based networks usually rely heavily on the entire shape to classify. Still,  certain important parts can be found so that once distorted, the score will drop quickly. Besides, symmetry is very important for the networks to recognize certain classes. We believe that these results improve our understanding of those networks and may help improving their design in the future.

\section{Related Work}
\noindent \textbf{Classifier visualization}
Using heatmaps to visualize networks has attracted much research effort these years. There are two main categories of approaches: gradient-based and perturbation-based. Gradient-based approaches utilizes the gradients of the output score w.r.t. the input as the standard of measuring input contribution~\cite{saliencymap,deconvnet,gbp,lrp,deeplift,ig}. Perturbation-based methods, on the other hand, perturb the input and examine which parts of the input have the largest influence on the output. Object detectors in CNNs \cite{perturb}, Real Time Image Saliency \cite{mask2}, Meaningful Perturbation \cite{mask}, RISE \cite{deletion} and I-GOS \cite{igos}  belong to this family. 

As far as we know, \cite{saliency} is the only prior work we know that attempts to visualize point cloud networks. \cite{saliency} uses a gradient-based approach and calculates the gradients of the output score with respect to the straight line from median to the input points and regards those gradients as saliency. %However, their method only works well for networks that have a max-pooling layer, and otherwise requires retraining of the classifier to take into account the irregular shape they created, whereas our goal is to be applicable to any type of deep network model without re-training.

%A close relative of perturbation-based visualization is adversarial attack \cite{adv,adv2}, in that both tasks seek the smallest change to the input that drop the prediction score of a deep network significantly. The difference is that visualization tasks want the perturbed features to be natural whereas most adversarial attacks tries to create data that fall outside of training distribution. The recent perturbation method I-GOS\cite{igos} explicitly involved techniques to prevent a visualization perturbation from becoming adversarial.

\noindent \textbf{3-D shape morphology}\label{sec:3dmorph}
There has been active research in smoothing and fairing 3-D structures. For mesh smoothing, \cite{taubin2} has proposed a method based on diffusion, and proved it to serve as a low-pass filter and is anti-shrinkage. However, as \cite{tangent} pointed out, this diffusion method is flawed due to its unrealistic assumption about meshes. \cite{tangent} proposed a scheme based on curvature flow, where a local ``curvature normal" is computed at each vertex and the diffusion is based on it. Meshes are easier to smooth than point clouds because they provide readily estimated planes that can be used to compute curvature. Noise-removal schemes that directly operate on point clouds were proposed in \cite{pss} and \cite{pcsmoothing1}. Most of these methods are based on moving least-squares \cite{mls} with local plane/surface fitting. However, the goals of these approaches are mainly removing noises, rather than gradually morphing the shape to one with constant curvature as in our goal.

%In terms of mathematical morphology, several work aimed to extend well-known 2-D morphological operations to point clouds \cite{pointmorph,lien}. In \cite{pointmorph}, a point set surface is fitted for the point cloud to get a signed distance function (SDF) representation for the point cloud, and then a point structuring element (PSE), which is a SDF itself, is fitted for each point using mean shift. Finally, the morphological projection of the point can be computed using the PSE. \cite{lien} proposed a purely point-based approach for defining the Minkowski sum for point clouds, which is fast and simple. However, most of these techniques require the shape to be closed and orientable, i.e., have an ``inside" and an ``outside", both being assumptions we would not like to make in order to render our algorithm more generally applicable.

\section{Methods}
Throughout this paper we work on a point cloud with $N$ points, denoted as $\mathbf{P} = [p_1, \ldots, p_N]$, where $p_i \in \mathbb{R}^3$ is a 3-tuple of $x, y, z$ coordinates. Denote a neighborhood of $p_i$ as $\mathcal{N}(p_i)$ and $K$ as the size of the neighborhood. Let $\textrm{diag}(\cdot)$ represent the operator taking a vector and making it a diagonal matrix, $\mathbf{I}$ be the identity matrix, and $\mathbf{1}$ be the vector of all $1$s.
\subsection{Smoothing Point Clouds}
\label{sec:smooth}
Our goal is to smoothly morph a point cloud into a feature-less shape. We regard ``curvature" on the surface as features here, since edges and corners are all areas of large curvatures on the surface that are distinct from their surroundings. Hence, we want the curvature on the entire point cloud to be constant or has little variance. Assuming the underlying manifold is closed, this goal is equivalent to morphing the shape into a sphere.
%If the shape is already on a plane, then we aim to make the curvature on its boundary constant, i.e., smoothing it into a disk. The total number of points should stay the same, and each point should be traceable from its initial position to its final position. In the following we first describe the classical Taubin smoothing \cite{taubin2} on meshes, then our algorithm.
\subsubsection{Taubin Smoothing} 
Our idea is inspired by Taubin smoothing \cite{taubin2}, a classical technique for meshes. In Taubin smoothing, the local Laplacian at a vertex $p_i$ is linearly approximated using the umbrella operator:
\begin{equation}\label{eq:lap}
L(p_i)=\dfrac{1}{K}\sum_{j\in\mathcal{N}(p_i)}(p_j-p_i).
\end{equation}
This approximation assumes unit-length edges and equal angles between two adjacent edges around a vertex \cite{tangent}. $L(p_i)$ has a matrix form $
L(\mathbf{P}) = - \mathbf{L} \mathbf{P}$
where $\mathbf{L} = \mathbf{D} - \mathbf{A}$ is the Laplacian matrix, assuming $\mathbf{A}$ is the $K$-nearest neighbor graph adjacency matrix in $\mathbf{P}$ and $\mathbf{D} = \textrm{diag}(\mathbf{A} \mathbf{1})$ is the diagonal degree matrix of each point (here $\mathbf{A} \mathbf{1}$ means the matrix multiplication between $\mathbf{A}$ and an all-one matrix. $\mathbf{A1}$ has constant $K$s on its diagonal). 
%, so that the discrete second derivative at $p_i$ on any direction $\vec{u}$ can be defined as:
%\begin{equation}\label{eq:dislap}
%L_{\vec{u}}(p_i)=\dfrac{1}{2}(p_{i+1}(\vec{u})-p_i)-\dfrac{1}{2}(p_{i}-p_{i-1}(\vec{u})),
%\end{equation}
%supposing $p_{i-1}$ and $p_{i+1}$ are the points right before and after $p_i$ along the direction $\vec{u}$. Usually $\vec{u}$ is chosen along the line from a mesh vertex to one of its neighbors.
%Say a point on a mesh has four neighbors. Since we are assuming equal angles between adjacent edges, ther are two pairs of neighbors that are colinear with each other and the center point. Then $\vec{u}$ here is just the direction of the two lines. 
%The umbrella operator sums up the second derivatives in all different  directions. 
Each vertex is then updated using the following scheme,
\begin{equation}\label{eq:taubin}
p_i'=p_i+\lambda L(p_i), 
p_i''=p_i'-\mu L(p_i')
\end{equation}
where $0<\lambda<1$ and $\lambda<\mu$. The first equation in Eq.(\ref{eq:taubin}) refers to a diffusion operator equivalent to $\mathbf{P} = (\mathbf{I} - \lambda \mathbf{L}) \mathbf{P}$, so that once this operation is carried out multiple times, most of the eigenvalues of $\mathbf{L}$ become close to zero and henceforth the points become more evenly distributed. 
%However, if only this is carried out, then the point cloud will ultimately collapse to a single mean. 
Furthermore, \cite{taubin2} proposed to add a step to prevent shrinkage, so that the volume enclosed by the underlying manifold does not decrease. An intuition behind Taubin smoothing is that the first equation in Eq. (\ref{eq:taubin}) attenuates the high frequencies and the second one magnifies the remaining low frequencies.
%, thus preventing shrinkage. However, as \cite{tangent} pointed out, this diffusion method is flawed due to its unrealistic assumption about meshes. 

\begin{figure*}[htb]
\begin{subfigure}{.48\linewidth}
\includegraphics[width=0.95\linewidth]{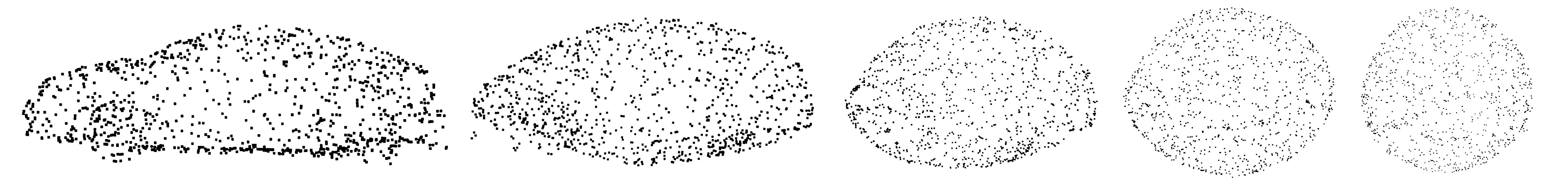}
\captionof{figure}{3-D version of our algorithm on a car}
\label{demo_car}
\end{subfigure}\hfill
\begin{subfigure}{.48\linewidth}
\includegraphics[width=0.95\linewidth]{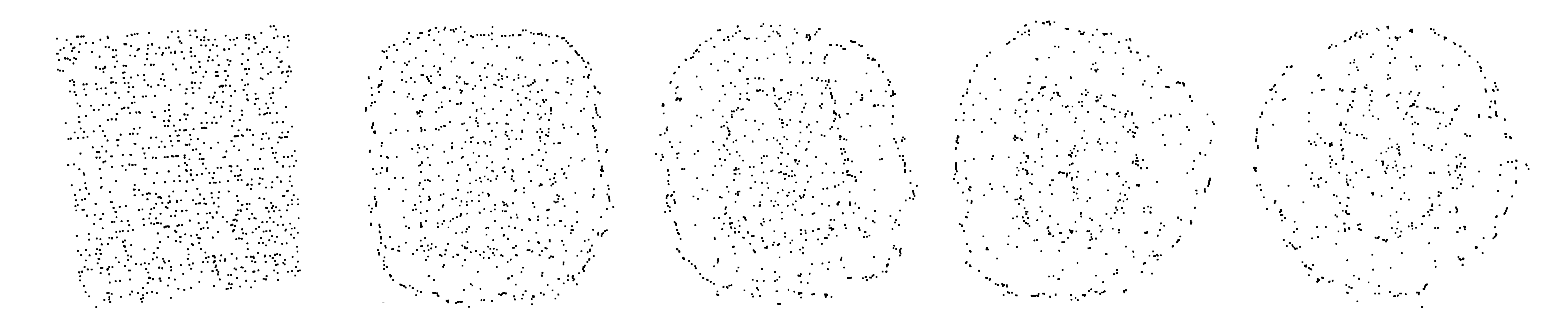}
\captionof{figure}{2-D version of our algorithm on a curtain}
\label{demo_curtain}
\end{subfigure}
\vspace{0.1in}
\caption{Demonstrations of our smoothing algorithm on two shapes from ModelNet40.}
\vskip -0.15in
\label{demo}
\end{figure*}

\subsubsection{Our algorithm}
\label{sec:algo}

\begin{figure*}[htb]
\begin{subfigure}{.5\linewidth}
\includegraphics[width=0.95\linewidth]{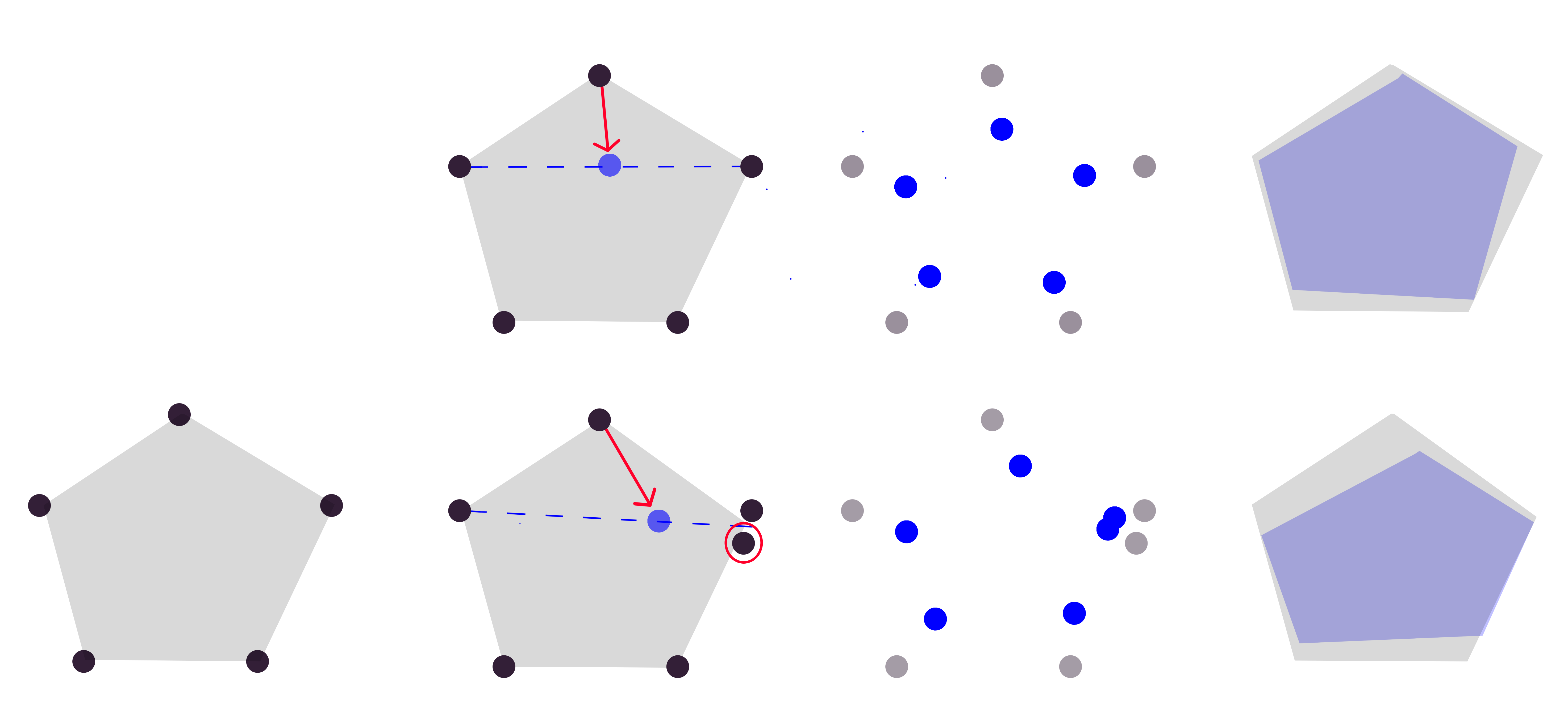}
\captionof{figure}{Laplacian smoothing. The resulting shape is distorted by the addition of one single noisy point (the second row).}
\label{average}
\end{subfigure}\hfill
\begin{subfigure}{.43\linewidth}
\includegraphics[width=0.95\linewidth]{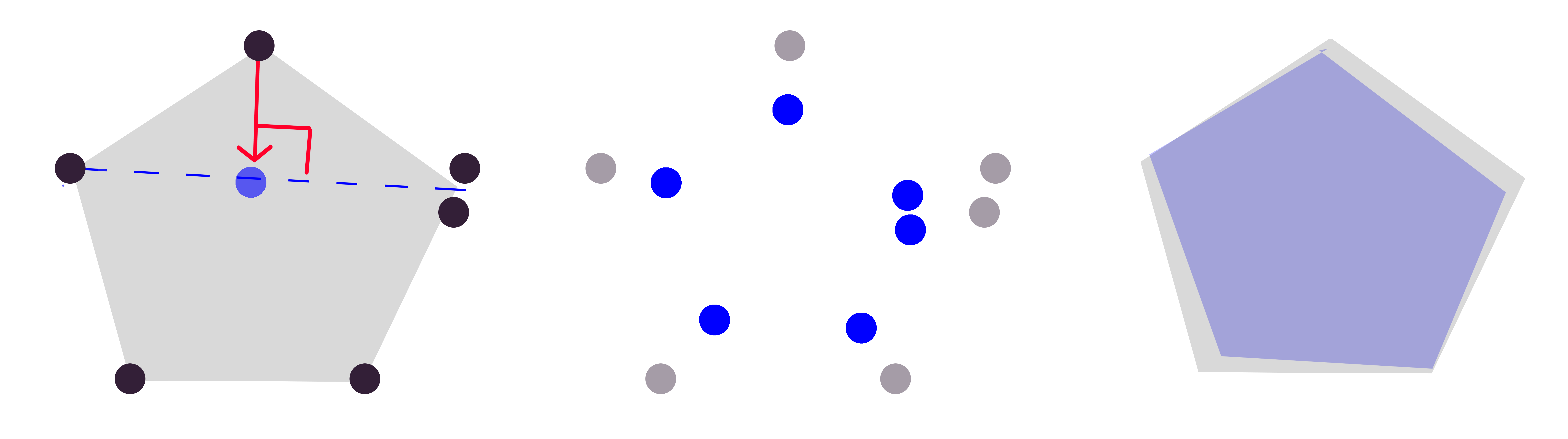}
\captionof{figure}{Curvature normal smoothing based on plane fitting. If we use a locally fitted plane to update the position of the points, then the resulting shape does not distort due to uneven distribution of the points.}
\label{plane}
\end{subfigure}
\vspace{0.1in}
\caption{Comparison between (a) Laplacian smoothing  and (b) the proposed curvature normal smoothing based on plane fitting. Left: original shape; Middle: smoothing results on the points; Right: Comparison of the underlying shapes of the original and new point set.}
\label{fivepointcomparison}
\vskip -0.1in
\end{figure*}
Based on the above diffusion formulation and with suitable parameter choices, Taubin smoothing should be able to smooth using any self-adjoint compact operator beyond the Laplacian operator \cite{zhu2007operator}. \cite{tangent} as an example smooths on the curvature normal operator on meshes. %However, their curvature approximation requires the data in the form of a mesh, which is different from a point cloud. 
In this paper, we approximate the mean curvature at a point by calculating its distance to a plane locally fitted to its neighborhood. Fitting such a plane allows us to be more robust to noisy input point clouds (Fig. \ref{fivepointcomparison}). Afterwards, we gradually filter out high frequency changes in curvature on the underlying surface of the point cloud.
If the underlying shape is a closed manifold, our algorithm will be able to smooth it approximately into a sphere, where curvature is constant everywhere. %Additionally, our algorithm leaves the density distribution of the point cloud unchanged as shown in sec. \ref{sec:expsmooth}, i.e., the distances between points remain roughly the same before and after morphing.

To fit a local plane $H=\{x:\langle x,\vn\rangle+D=0,x\in\mathbb{R}^3\},\vn\in\mathbb{R}^3,||\vn||=1$ for each point $p_i$, we minimize the least-squares error:
\begin{equation}\label{eq:localplane}
\argmin_{\vn,D} \sum_{j\in \mathcal{N}(p_i)} \left( \langle p_j,\vn\rangle +D\right)^2
\end{equation}

Let $h_i$ denote the position of $p_i$ after being projected onto $H$ (i.e. $h_i=p_i-(\langle p_i,\vn\rangle+D)\cdot \vn$). Then $h_i-p_i$ is the vector pointing from the point $p_i$ to the plane $H$. Note that the direction of $h_i$ is just the surface normal at $p_i$. However, we hold that the distance to the plane is an approximation to the mean curvature, and coincides with the curvature under some simplifying assumption.

\begin{theorem}
Let $p_i\in \mathbb{R}^3$ be a point in point cloud. Let $H=\{x:\langle x,\vn\rangle+D=0,x\in\mathbb{R}^3\},\vn\in\mathbb{R}^3,||\vn||=1$ be the plane fitted to the neighbors of $p_i$. Let $h_i$ be the projection of $p_i$ on $H$. Assuming $p_i$'s neighbors distribute evenly and densely on a ring surrounding $h_i$, then the curvature normal at $p_i$ can be approximated by the expression $\dfrac{1}{2k^2}(h_i-p_i)$, where $k$ is the distance from $p_i$ to any of its neighbor.
\end{theorem}
 See the supplementary material for the proof. %stating that the curvature normal at a point $p_i$ can be approximated by the expression $\dfrac{1}{2k^2}(h_i-p_i)$, where $k$ is the distance from $p_i$ to any of its neighbor.

With that result, we can accommodate the smoothing algorithm from \cite{taubin2} as follows:
\begin{equation}\label{eq:potato}
p_i'=p_i+\lambda\left(h_i -p_i\right),
p_i''=p_i'-\mu\left(h_i'-p_i'\right)
\end{equation}
where $0<\lambda<1$, $\lambda<\mu$ and $h_i'$ refers to the projection of $p_i'$ on a new plane $H'$ fitted for $p_i'$. Thus instead of moving the point toward the mean of its neighbors, we move it directly toward the locally fitted plane. We call the first equation in Eq. \ref{eq:potato} the ``erosion" round, and the second one  the ``dilation" round.

To deal with degenerate cases where the point cloud is already on a plane, we further extend the algorithm to a 2-D case (Fig. \ref{demo_curtain}). Here the goal is to filter out high frequency changes in curvature on the boundary, transforming the plane to a disk. In this case, assuming all the neighborhood points $\mathcal{N}(p_i)$ are on the plane, we fit a line
$H'=\{x:\langle x,\vn'\rangle+C=0,x\in\mathbb{R}^2\},\vn'\in\mathbb{R}^2,||\vn'||=1$ for $w_i=(0,0)$ by minimizing the least-squares error:
\begin{equation}\label{eq:localline}
\argmin_{\vn',C} \sum_{j\in \mathcal{N}(p_i)} \left( \langle p_j,\vn'\rangle +C\right)^2
\end{equation}
where each $w_j$ is the projection of $p_j$ to the plane $(\vec{u}, \vec{v})$. Let $q_i$ be the projection of $p_i$ on line $H'$. We update $p_i$ in the same fashion as in the 3-D case:
\begin{equation}\label{eq:line}
w_i'=w_i+\lambda\left(q_i-w_i\right), 
w_i''=w_i'-\mu\left(q_i'-w_i'\right)
\end{equation}
Denote the final 2D coordinates as $w_T=(u_T,v_T)$, we convert it back to 3-D by calculating $p_i'=p_i+u_T\vec{u}+v_T\vec{v}$. In reality, due to noises, many points are not exactly on a plane. We project them to their local planes $H$ first, and then calculate the $uv$-coordinates from their projected location $h_i$. Note that we still shift the point from its original location $p_i$, not its projected location $h_i$.
In actual implementation, the 2-D version is used together with the 3-D version and is always run first.
%\todo{what happens if the object is not on a plane?} 
For example, in an ``erosion" round, we run the first equation in Eq. (\ref{eq:line}), then the first equation in Eq. (\ref{eq:potato}); in a ``dilation" round, we run the second equation in Eq. (\ref{eq:line}), then the second equation in Eq. (\ref{eq:potato}). Empirically this seems to generalize well on both planar and non-planar surfaces, we believe the reason is that on non-planar surfaces the line fitting usually falls close to the point itself, hence the planar version hardly moves any point at all. By utilizing both of them at every iteration, we avoid introducing an extra threshold to decide whether a neighborhood is on a plane.

\subsection{Visualizing Point Cloud Classifiers}\label{sec:igos}
Our goal is to find the most important points that decide the output of a classifier. Following the idea of ``mask" from \cite{mask}, we achieve this goal by finding such a mask that the classification score is minimized when the mask is applied to the point cloud, and the score is maximized when the reverse of the mask is applied. Inspired by \cite{ig} and \cite{igos}, we use an \textit{integrated} loss to train our mask.

Let mask $\mathbf{M}$ be of the same size as the point cloud $\mathbf{P}$, initialized with all zeros. Mask values are always between $[0,1]$, where $0$ means no smoothing and $1$ means fully smoothing.  Let our baseline point cloud $\mathbf{P}_0$ be the fully smoothed point cloud (e.g. sphere) and let our baseline mask be $\mathbf{M}_0 = \mathbf{11}^\top$, so that when applied to the shape, the shape becomes $\mathbf{P}_0$. The idea of an \textit{integrated} mask is that we gradually morph $\mathbf{M}$ to $\mathbf{M}_0$, which is a global minimum for the classification score loss, and collect the classification score loss along the path:
\begin{equation}\label{eq:del}
L_{del}=\int_{\alpha=0}^1 f_c(\Phi(\mathbf{P},\mathbf{M}+\alpha (\mathbf{M}_0-\mathbf{M}))) d\alpha
\end{equation}
and
\begin{equation}\label{eq:ins}
L_{ins}=-\int_{\alpha=0}^1 f_c(\Phi(\mathbf{P},\overline{\mathbf{M}}+\alpha (\overline{\mathbf{M}_0}-\overline{\mathbf{M}}))) d\alpha
\end{equation}
where $f_c(\cdot)$ represents the classifier on the class $c$, $\overline{\mathbf{M}}\equiv \mathbf{1}-\mathbf{M}$ denotes the reverse of the mask and $\Phi$ represents the action of applying the mask to the point cloud. $L_{del}$ indicates the classification score should plunge as crucial features are gradually deleted ($\mathbf{P}$ to $\mathbf{P}_0$) and $L_{ins}$ indicates the classification score should increase significantly as crucial features are gradually inserted. The benefit of integrated gradients is that they are more likely pointing to a global optimum for the unconstrained problem of only minimizing the classification loss of a single mask, so that the optimization can evade local optima and achieve better performance. In practice, we approximate the integration process in the above equations by dividing it into 20 steps and average through the 20 losses.

However, with classification loss only, the algorithm might as well return the baseline mask $\mathbf{M}_0$. In order to identify the most important set of points, we must constrain the sum of mask values by using an $l1$ loss $L_{l1}=\dfrac{1}{N}||\mathbf{M}||_1$.
%We would also like the mask to be smooth by adding a \textit{total-variation} loss $L_{tv}=\dfrac{1}{N}\sum_{\mathbf{M}}\dfrac{1}{K} \sum_{j\in \mathcal{N}(p_i)} |m_j-m_i|$.

Altogether, our mask is trained using the following losses
\begin{equation}\label{eq:loss}
\min_{\mathbf{M}} L_{del} + L_{ins} + \lambda_{l1} L_{l1}(\mathbf{M})
\end{equation}

One difficulty of this algorithm is how to implement $\Phi(\cdot)$ as a differentiable masking operation. In 2-D images, we can simply use a weighted (by $m_i$) average of the actual pixel value and the baseline pixel value. However, in point clouds, if we directly push a point toward its corresponding baseline position, undesirable (out-of-distribution) sharp structure might appear. 

Ideally, we want to run more smoothing iterations on points with higher mask value. Unfortunately, the smoothing process is not parametrized by mask values.

In practice, we construct a differentiable $\Phi(\cdot)$ by precomputing 10 intermediate shapes with increasing level of smoothness. Since the smoothing method we introduced is iterative, we simply run the algorithm for $10S$ iterations and capture the shape after each $S$ iterations.
We approximate the ideal mask smoothing operation by combining the 10 shapes:
\begin{equation}\label{eq:mask}
\Phi(p_i,m_i)=\frac{\sum_{l=0}^{10}\exp(- \alpha \|10\cdot m_i - l\|^2)p_{i,l}}{\sum_{l=0}^{10}\exp(- \alpha \|10\cdot m_i - l\|^2)}
\end{equation}
where $p_i$ is a point with a mask value $m_i \in [0,1]$, $l$ refers to the $l$-th point cloud in our sequence of precomputed smoothed shapes ($l=10$ refers to $\mathbf{P}_0$ and $l=0$ refers to the original shape), and $p_{i,l}$ refers to the position of the $i$-th point in the $l$-th point cloud. Here, we are using a Gaussian kernel to assign weights to each level of the masks. The closer $10\cdot m$ and $l$, the higher the weight. For example, when the mask value at $p_i$ is nearly transparent, $m$ will be low, and thus masks with lower smoothing level $l$ will gain greater weights. After obtaining the masked shape, we apply the point cloud classifier to get the classification score for the losses, and then calculate the gradients.

Under our algorithm, the mask converges quickly (we typically only need 30 optimization steps for each shape), and the resulting masks only make small changes to the original point clouds with a large impact on the prediction score, and are interpretable by human (as shown in Fig. \ref{fig1}). Finally, we output the mask as our saliency map.

\section{Experiments}
\label{sec:exp}
We have conducted two types of experiments. First, we compare our smoothing algorithm against several baselines to validate its  smoothing capability. Second, we visualize point cloud classifiers using PCI-GOS, compared it with baselines as well as another visualization technique proposed by \cite{saliency}, and performed several ablation studies. All experiments are conducted on the test split of the ModelNet40 dataset, with the classifier trained on the training split. Each shape contains 1024 randomly sampled points, and only $xyz$ location information.
Parameters of our smoothing algorithm are: $\lambda=0.7,\mu=1.0$, $K$ grows from 20 to 60. 
We run 80 iterations on each shape (one iteration $=$ one ``erosion"  $+$ one ``dilation"). 

%\todo{a few more details about the ModelNet40 dataset please}.
\subsection{Point cloud smoothing}
\label{sec:expsmooth}
%Neighborhood has an order of 2 (meaning that $\mathcal{N}(p_i)=\mathcal{N}_k(p_i)\cup \bigcup_{j\in \mathcal{N}_k(p_i)} \mathcal{N}_k(p_j)$ where $\mathcal{N}_k(\cdot)$ is the $k$-nearest-neighbor operator). 

Since there were few prior work that aim at morphing point clouds into spheres, we compare against several other plausible baselines. First note that directly applying Gaussian blur to point coordinates is not a valid baseline, because Gaussian blur tends to smooth the coordinate values, which results in pushing neighboring points to all have the same coordinates, leading to a skeleton effect. We compare against three baselines:

\textbf{Meshing, then smoothing.} This idea converts the point cloud to a mesh and then applies mesh-based smoothing techniques such as \cite{tangent} to the result. For our goals, we chose \cite{gp3} as an algorithm that does not change the number of points and maintains a 1-1 correspondence with the original point cloud. Due to the noisiness and sparsity of the point cloud, the meshing result is often not ideal. 

\textbf{Directly applying mesh smoothing techniques to points.} Instead of explicit meshing, we construct an \textit{implicit mesh} by assuming a point is connected to all its neighbors. Then, we directly apply mesh smoothing techniques to the point cloud. However, the uneven distribution of points in a point cloud quite often distorts the result.%, in a method such as Taubin smoothing. %Though \cite{fuji} and \cite{tangent} have proposed improvements for irregular meshes, they explicitly exploit edge information, which is not available in point cloud data (and requires explicit meshing as above).

\textbf{Fitting a quadratic surface.} We fit a quadratic surface to the local neighborhood instead of a plane. A quadratic surface allows analytic computation of the curvature, which is in principle a better approximation than the plane. We implemented the closed-form quadratic fitting algorithm following \cite{quadric}. However, quadratic surfaces have a large degree of freedom and thus even a tiny noise can render an overfitting quadratic type or direction. %As illustrated in Fig. \ref{quadric}, the border of the car shape ends up consisting of quadratic lines curving outward instead of inward. 
%\todo{get rid of Fig. 3, put Fig. 5 and Fig. 6 on the same row please} %Unfortunately, 3-D point cloud data are usually full of noises. 

\begin{table*}[htb]
\begin{small}
\caption{\small Comparison of point cloud smoothing algorithms. Mesh refers to meshing and smoothing. Taubin refers to directly applying Taubin smoothing to point clouds. Only our algorithm succeeds in both removing features from the surface and keeping the morphing process smooth.  For $l=0$ (initial shapes), CSD=0.10, MR=0.83. For CSD, lower is better; for MR and DDS, higher is better.}
\label{tb:eva1}
\begin{center}
\begin{tabular}{c|c|cccccccccc}
%Algorithm & Metric &&&&&&&&&&\\
\hline
\multicolumn{2}{c|}{Smooth level}  & 1 & 2 & 3 & 4 & 5 & 6 & 7 & 8 & 9 & 10\\\hline

\multirow{3}{*}{\shortstack{Mesh}} 
& CSD & 0.10 & 0.10 & 0.11 & 0.12 & 0.13 & 0.14 & 0.16 & 0.17 & 0.19 & 0.20\\
& MR & 0.82 & 0.82 & 0.82 & 0.82 & 0.82 & 0.83 & 0.86 & 0.85 & 0.83 & 0.83 \\
%& DDS1 & 0.12 & 0.12 & 0.05 & \textit{0.02} & \textit{0.04} & 0.28 & 0.57 & 0.63 & 0.63 & 0.63\\
& DDS & 0.40 & 0.67 & 0.62 & 0.63 & 0.60 & 0.58 & 0.48 & 0.38 & 0.40 & 0.30\\\hline
\multirow{3}{*}{\shortstack{Taubin}}
& CSD & 0.10 & 0.11 & 0.11 &  0.11 & 0.11 & 0.11 & 0.10 & 0.10 & 0.09 & 0.09\\
& MR  & 0.83 & 0.84 & 0.83 & 0.87 & 0.88 & 0.86 & 0.86 & 0.86 & 0.75 & 0.73 \\
%& DDS1 & 0.28 & 0.08 & \textit{0.03} & \textit{0.00} & \textit{0.00} & \textit{0.00} & \textit{0.00} & \textit{0.02} & 0.24 & 0.60 \\
& DDS & 0.90 & 0.92 & 0.74 & 0.87 & 0.83 & 0.81 & 0.69 & 0.74 & 0.43 & 0.66\\\hline
\multirow{3}{*}{\shortstack{Quad}} 
& CSD & 0.11 & 0.12 & 0.12 & 0.12 & 0.12 & 0.12 & 0.13 & 0.13 & 0.13 & 0.13\\
& MR  & 0.79 & 0.80 & 0.81 & 0.81 & 0.83 & 0.83 & 0.83 & 0.84 & 0.83 & 0.83\\
%& DDS1 & 0.38 & 0.64 & 0.77 & 0.86 & 0.91 & 0.94 & 0.96 & 0.98 & 0.97 & 0.97 \\
& DDS & 0.76 & 0.83 & 0.84 & 0.89 & 0.82 & 0.89 & 0.92 & 0.92 & 0.88 & \textbf{0.94}\\\hline
\multirow{3}{*}{Ours} 
& CSD & 0.08 & 0.07 & 0.07 & 0.06 & 0.06 & 0.06 & 0.06 & 0.06  & 0.06 & \textbf{0.05}\\
& MR & 0.85 & 0.87 & 0.88 & 0.89 & 0.91 & 0.92 & 0.94 & 0.94 & 0.95 & \textbf{0.95}\\
%&  DDS1 & 0.09 & 0.29 & 0.10 & 0.26 & 0.11 & 0.20 & 0.09 & 0.13 & 0.05 & 0.07\\
& DDS & 0.60 & 0.75 & 0.68 & 0.72 & 0.64 & 0.66 & 0.59 & 0.60 & 0.56 & 0.58\\
\hline
\end{tabular}
\end{center}
%\vskip -0.15in
\end{small}
\end{table*}

\begin{table*}[h]
%\vskip -0.1in
\begin{small}
\caption{\small PCI-GOS compared to other methods using the \textit{deletion} and \textit{insertion} metrics (averaged over 40 classes), conducted with the PointConv classifier. We evaluate the scores using both Point Deletion/Insertion (directly remove/add points from the point cloud) and Curvature Deletion/Insertion (move points using our curvature-based smoothing). For deletion, lower is better, for insertion, higher is better}
%\vskip -0.2in
\label{tb:baseline}
\begin{center}
\begin{tabular}{c||c|c|c|c||c|c}
\hline
& mask-only & ig-only & Zheng et al. & Ours & Zheng et al.\cite{saliency} & Ours\\\hline
& \multicolumn{4}{c||}{Curvature Del./Ins.} & \multicolumn{2}{c}{Point Del./Ins.} \\
\hline
\textit{Deletion}$\downarrow$  &0.2514 &0.2812&0.2597&\textbf{0.2214}&\textbf{0.2793}&0.4073\\
\textit{Insertion}$\uparrow$ &0.2917 &0.3970 &0.4219&\textbf{0.4502}&0.4976&\textbf{0.5215}\\
%Difference$\uparrow$ &0.0403 &0.1158 &\underline{0.2183}&0.1623&0.1142&\textbf{0.2287}\\
\hline
\end{tabular}
\end{center}
%\vskip -0.2in
\end{small}
\end{table*}

\begin{table*}[h]
\begin{small}
%vskip -0.15in
\caption{\small Ablation study for $l1$-loss, $ins$-loss and mask size using \textit{deletion} and \textit{insertion} metrics. As shown, all losses are necessary for maximizing the performance of the algorithm.}
%\vskip -0.15in
\label{tb:ablation}
\begin{center}
\begin{tabular}{c||c|c|c|c}
\hline
&  w/o $l1$ &  w/o $ins$ & msize=1024 & full \\\hline
\textit{Deletion}$\downarrow$ & 0.2226  & \textbf{0.1965} & 0.2463 & 0.2214\\
\textit{Insertion}$\uparrow$  & 0.4419 &  0.3610 & 0.4109 & \textbf{0.4502}\\
%Difference$\uparrow$ & 0.2193 & 0.1645 & 0.1646 & \textbf{0.2287}\\
\hline
\end{tabular}
%\vskip -0.3in
\end{center}
\end{small}
\end{table*}

For a quantitative comparison against these baselines, we propose \textbf{three} metrics to evaluate our smoothing algorithm: \textit{curvature standard deviation} (CSD), \textit{min-max ratio} (MR) and \textit{density distribution similarity} (DDS). The first two ensure that the final shape is feature-less as desired, and the last one ensures that the morphing process does not bring abrupt changes to the point cloud. Please refer to supplementary materials for more explanation about these metrics. Ten intermediate point clouds with increasing level of blurriness are sampled. 
%The metrics are calculated for all of them and the averaged results across the entire ModelNet40 testing set are listed in Table \ref{tb:eva1}. 

From the experiment results, all baseline algorithms fail to eliminate large curvatures on the surface. %Their curvature variance has even increased for some blurred levels.
All of them fail to improve MR at all, which means the final shape is not sphere-like as desired. Only our algorithm succeeds in both removing features from the surface and keeping the morphing process smooth.

\begin{figure*}[htb]
\begin{subfigure}{.48\linewidth}
\includegraphics[width=0.95\linewidth]{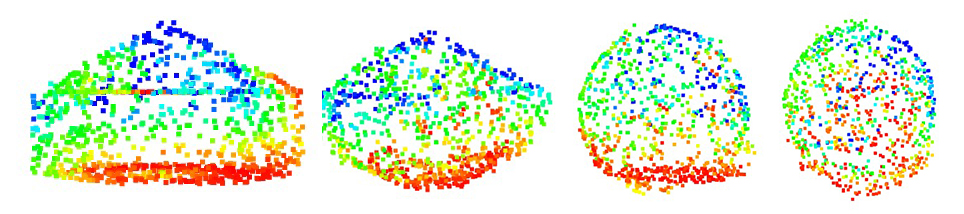}
\captionof{figure}{Tent: 1.00, 0.08, 0.89, 0.89 (top).}
\end{subfigure}\hfill
\begin{subfigure}{0.48\linewidth}
\includegraphics[width=0.95\linewidth]{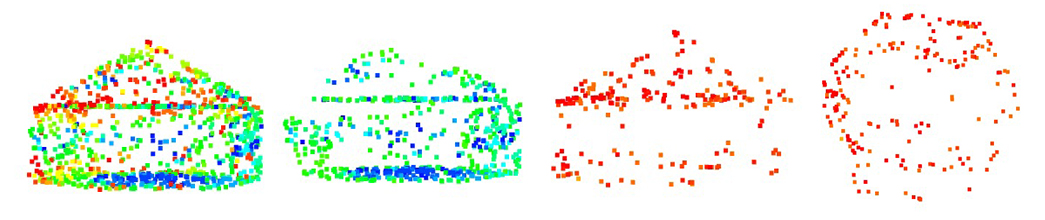}
\captionof{figure}{Tent: 1.00, 0.00, 0.89, 0.89 (top).}
\end{subfigure}
\vspace{0.1in}
\caption{\small (a) Results of our algorithm; (b) Results of \cite{saliency}. From left to right: Original Image; The first deletion image with predicted confidence lower than $0.1$; The first insertion image with predicted confidence higher than $0.75$; Top-view of the third Image. The numbers indicate the respective predicted confidence (Best viewed in Color)}
\label{demo_tent}
\vskip -0.15in
\end{figure*}

\begin{figure*}[htb]
\begin{subfigure}{.20\linewidth}
\includegraphics[width=0.95\linewidth]{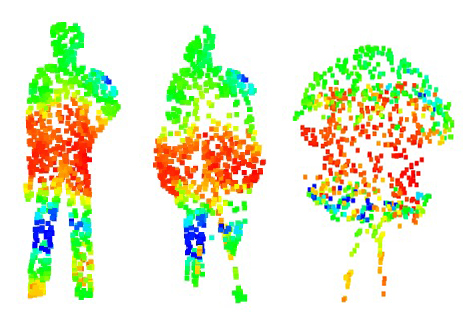}
\captionof{figure}{1.00, 0.05, 0.78.}
\label{person1}
\end{subfigure}
\begin{subfigure}{0.20\linewidth}
\includegraphics[width=0.95\linewidth]{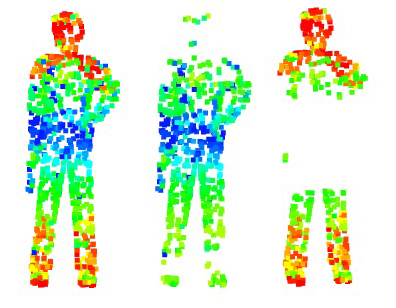}
\captionof{figure}{1.00, 0.11, 0.76.}
\label{person2}
\end{subfigure}
\begin{subfigure}{.29\linewidth}
\includegraphics[width=0.95\linewidth]{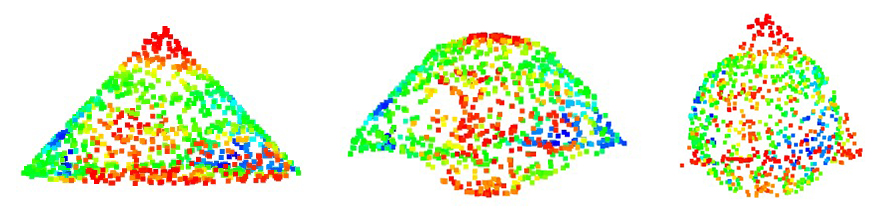}
\captionof{figure}{1.00, 0.02, 0.95.}
\end{subfigure}
\label{cone1}
\begin{subfigure}{0.29\linewidth}
\includegraphics[width=0.95\linewidth]{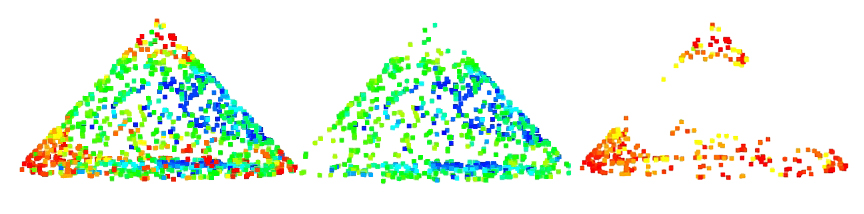}
\captionof{figure}{1.00, 0.00, 1.00.}
\label{cone2}
\end{subfigure}
\vspace{0.1in}
\caption{\small (a) (c) Results of our algorithm; (b)(d) Results of \cite{saliency}. From left to right: Original Image; The first deletion image with predicted confidence lower than $0.1$; The first insertion image with predicted confidence higher than $0.75$. The numbers indicate the respective predicted confidence (Best viewed in Color)}
\label{demo_diff}
%\vskip -0.15in
\end{figure*}

\begin{figure*}[htb]
\begin{subfigure}{.55\linewidth}
\includegraphics[width=0.95\linewidth]{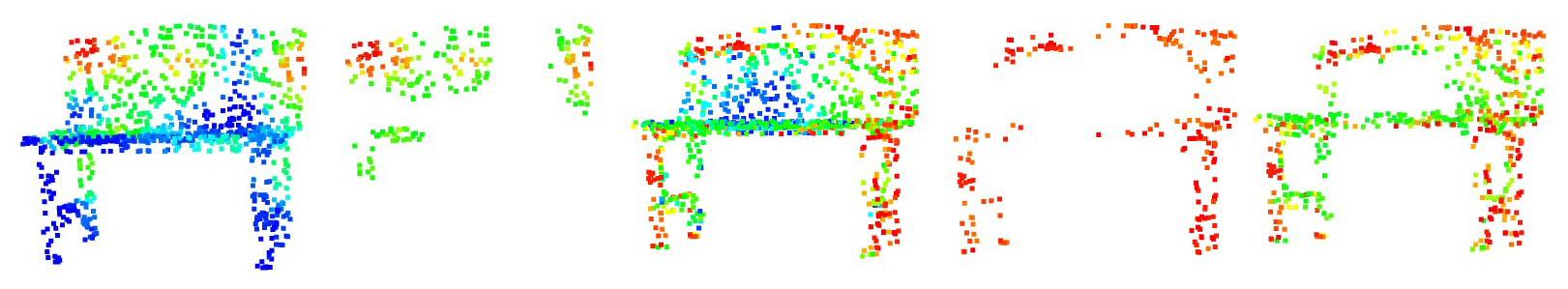}
\captionof{figure}{Ours: 0.86, 1.00[20]. \cite{saliency}: 0.86, 0.00[20], 0.49[60].}
\label{bench}
\end{subfigure}
\begin{subfigure}{0.43\linewidth}
\includegraphics[width=0.95\linewidth]{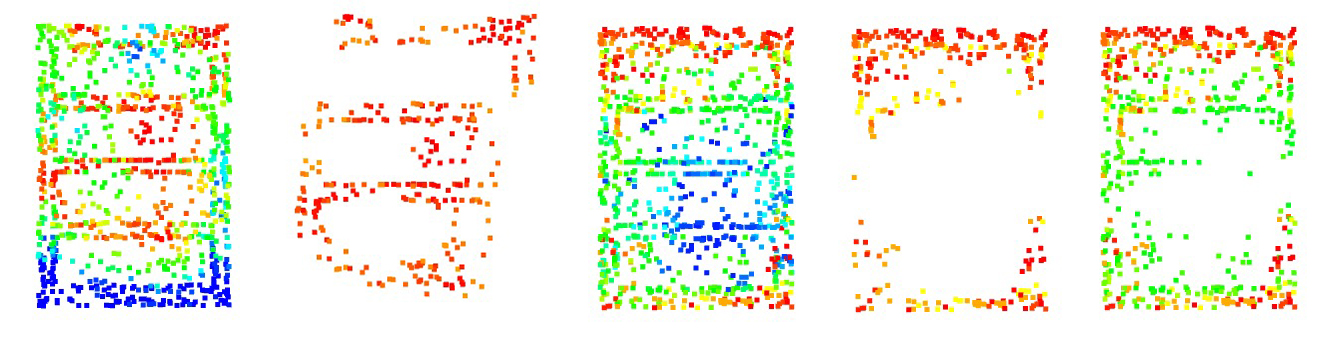}
\captionof{figure}{Ours: 0.99, 1.00[30]. \cite{saliency}: 0.99, 0.00[30], 0.91[60].}
\label{bookshelf}
\end{subfigure}
\vspace{0.1in}
\caption{\small Score[Insertion percentage] for a bench and a  bookshelf. Our highlighted points give rise to score more quickly than \cite{saliency} (Best viewed in Color)}
\label{demo_ins}
%\vskip -0.15in
\end{figure*}

\subsection{Classifier visualization}
\label{sec:expvis}

We experiment our PCI-GOS algorithm on PointConv \cite{pointconv}, a state-of-the-art point cloud classifier, with the ModelNet 40 test set. We use the \textit{deletion} and \textit{insertion} metrics proposed by \cite{deletion} to evaluate the heatmaps. Numbers displayed in the tables are the average scores along the deletion / insertion curves. Instead of point deletion / insertion, we use curvature deletion / insertion to evaluation our method. To delete top 5\% curvature means smoothing only the top 5\% points, and vice versa for insertion.
The color scheme used for saliency map in picture illustrations: blue (0.0) $\rightarrow$ green $\rightarrow$ red (1.0).

Table \ref{tb:baseline} lists results of our algorithm compared to several baselines and \cite{saliency}. Mask-only learns the mask using gradients instead of integrated gradients. Each mask goes through 300 iterations under this method, as opposed to 30 under PCI-GOS. Ig-only directly takes a one-step integrated gradient instead of an optimization process.

Our algorithm is optimized for curvature deletion/insertion, where curvature deletion means smoothing certain curved areas, and curvature insertion means smoothing all but those curved areas. It is shown that our approach outperforms both of these baselines. We also compare against Zheng \textit{et al.} \cite{saliency}. Here, note that the method in \cite{saliency}  is optimized for point deletion/insertion.To ensure fairness, we evaluate both methods on with both point and curvature del/ins. PC-IGOS and \cite{saliency} give similar performance when respectively using their own evaluation method, and perform worse when using each other's evaluation. As shown in Fig. \ref{demo_tent}, from our perspective, the most important feature for a tent is a flat ground, while from \cite{saliency}'s perspective, the most important features are the points along the skeleton. It is difficult to argue from visual results which one is better, but we believe this has provided different perspectives of the point cloud classifier.

Interestingly, PCI-GOS improves over Zheng \textit{et al.} \cite{saliency} on both insertion metrics. We hypothesize that this might be because our algorithm tends to highlight an entire surface rather than concentrate on the edge of a shape (see Fig. \ref{demo_ins}). E.g., in the case of bookshelf, ModelNet40 contains many classes that have similar skeleton, such as dresser, wardrobe, etc. Thus, a sole rectangular frame might not be able to help the classifier to make decision.

%Meanwhile, it is not surprising that our highlighted points do not drop the score quick when they are directly deleted. Given that they usually concentrate on a surface in order to morph the surface's curvature to a uniform sphere, deleting them will not influence the overall frame of the shape. Thus, the \textit{deleltion} curve might drop slowly.

Table \ref{tb:ablation} shows the ablation study for the $l1$-loss and the $insertion$-loss (Eq. \ref{eq:ins}). Without the $ins$-loss, the \textit{deletion} curve performs better and the \textit{insertion} curve worse as expected, since the algorithm now concentrates on looking for points that drop the score quickly but not necessarily give rise to the score quickly.
In practice, we also found a smaller mask size helps saliency learning. Usually, we train a mask size of 256 and upsample it to 1024 when applying in Equation~\ref{eq:mask}. Ablation study shows that directly optimizing a mask of 1024 points leads to worse results, perhaps because the additional points make the optimization problem harder to solve.

For class-wise deletion and insertion curves, please refer to our supplementary material. 

%One issue that is unique to point cloud smoothing is that, the fully-smoothed spherical shaped cloud may still have semantic meaning, different from 2D images where there is usually no information in a completely blurred image. In Table.~\ref{tb:mean_blur} we took $100$ random completely smoothed point clouds (spherical) and classified them with PointConv. The result showed significant confidence on  3 classes: vase, flowerpot and plant. This indicates that maybe the current visualizations will not be informative in these 3 classes. Alternative approaches to define ``featureless" or non-informative point clouds are interesting future work.

\section{Conclusions and Future Work}
In this paper, we propose a novel smoothing algorithm for morphing a point cloud into a shape with constant curvature, and PCI-GOS, a 3-D classifier visualization technique. 
%However, 
We regard the most important contribution of this paper to be a new direction for point cloud network visualization -- an optimization-based approach.
It is a bit difficult to compare our method and \cite{saliency} since the optimization goals are different. We generate quite different visualization results from prior work \cite{saliency}, but our insertion metrics are consistently higher than theirs, no matter evaluated using their methodology or ours. 
Additionally, our algorithm is more flexible with respect to learning goal. For example, by tuning up the coefficient of the $insertion$-loss, we can obtain a mask that tends to highlight points capable of giving rise to prediction score quickly. 
We hope the visualization results in this paper improve the understanding on those new
point cloud networks and we look forward to exploring better
definitions of ``non-informative" point clouds as well as
smoothing with features beyond curvature in future work.

\begin{figure*}[b]
\begin{center}
\begin{subfigure}{.32\linewidth}
\includegraphics[width=0.95\linewidth]{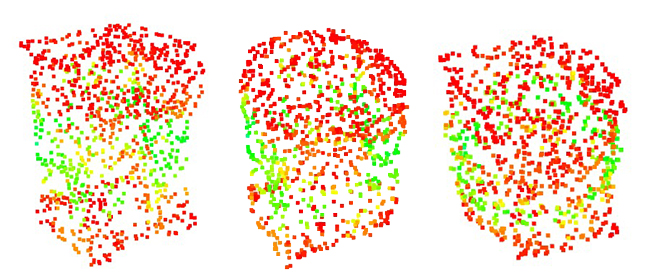}
\captionof{figure}{\footnotesize Nightstand:.56,.01(10),.82[60]}
\end{subfigure}
\begin{subfigure}{0.32\linewidth}
\includegraphics[width=0.95\linewidth]{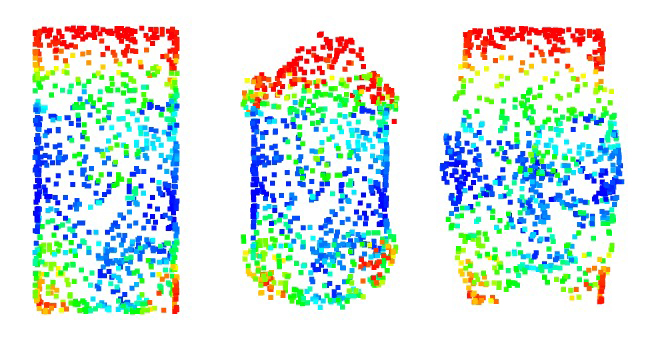}
\captionof{figure}{\footnotesize Wardrobe:.76,.00(10),.74[40]}
\end{subfigure}
\begin{subfigure}{0.32\linewidth}
\includegraphics[width=0.95\linewidth]{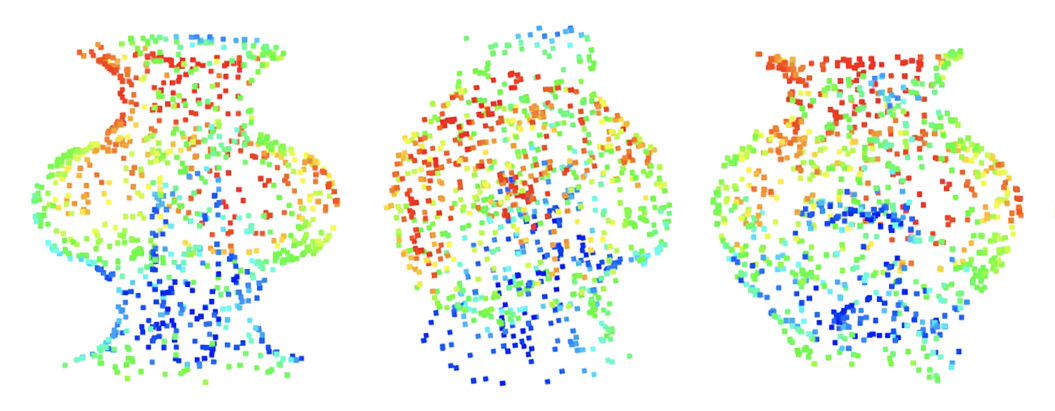}
\captionof{figure}{\footnotesize Vase:.83,.19(30),.82[70]}
\end{subfigure}

\begin{subfigure}{.32\linewidth}
\includegraphics[width=0.95\linewidth]{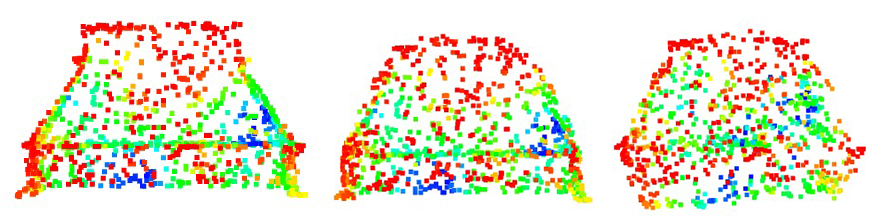}
\captionof{figure}{\footnotesize Rangehood:1,.10(10),.92[60]}
\end{subfigure}
\begin{subfigure}{0.32\linewidth}
\includegraphics[width=0.95\linewidth]{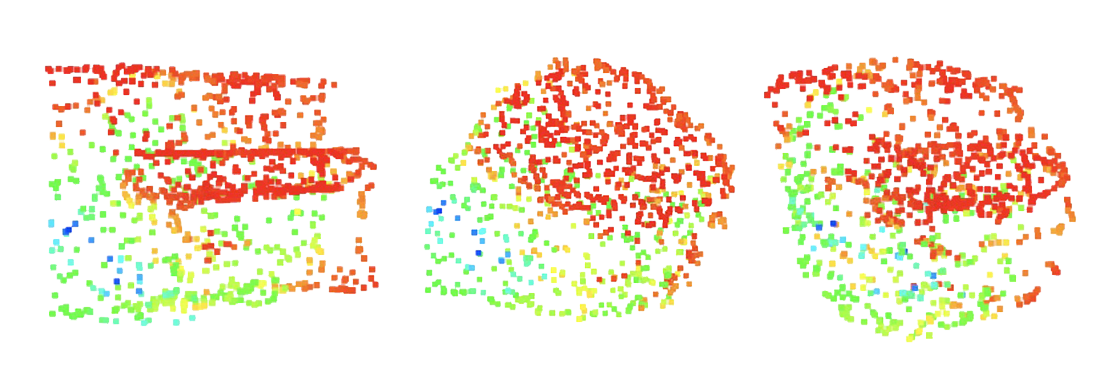}
\captionof{figure}{\footnotesize Piano:.99,.00(30),.79[70]}
\end{subfigure}
\begin{subfigure}{0.32\linewidth}
\includegraphics[width=0.95\linewidth]{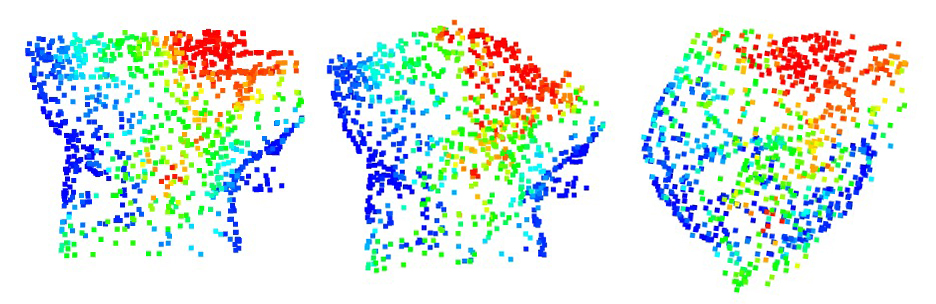}
\captionof{figure}{\footnotesize Toilet:1,.14(20),.83[40]}
\end{subfigure}
\vspace{0.1in}
\caption{\small More illustrations of our algorithm (leftmost the original shape). Class: Score(del\%)[ins\%].}
\label{more_demo}
\end{center}
%\vskip -0.15in
\end{figure*}

\section*{Acknowledgments}
This work was partially supported by the National Science Foundation (NSF) under Project \#1751402, USDA National Institute of Food and Agriculture (USDA-NIFA) under Award 2019-67019-29462, as well as by the Defense Advanced Research Projects Agency (DARPA) under Contract No. N66001-17-12-4030.

\clearpage
\onecolumn

\appendix
\appendixpage

\begin{figure*}[h]
\centering
\includegraphics[width=0.98\linewidth]{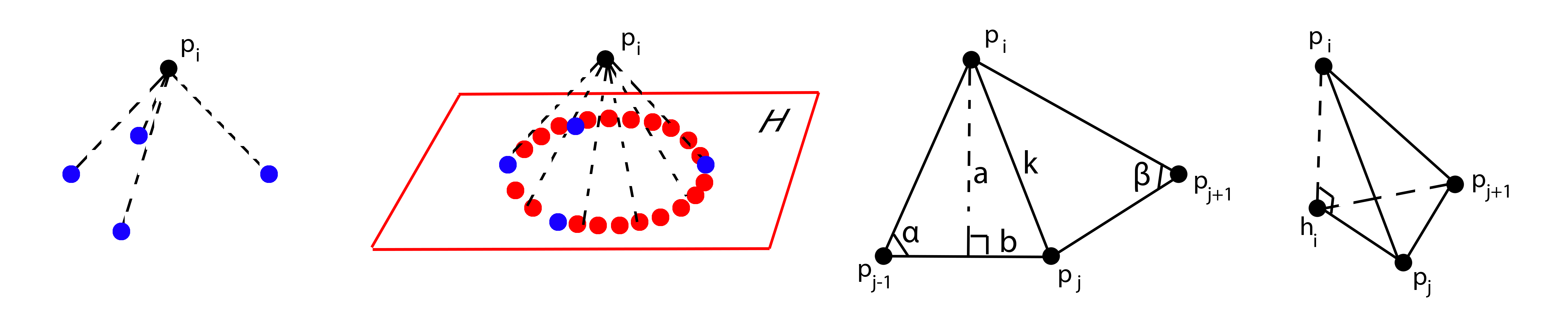}
\caption{Auxillary graph for proof in Appendix \ref{ap:proof}. From left to right: point $p_i$ and its actual neighbors (in blue), $p_i$ and its virtual neighbors (in red) and the fitted local plane $H$, enlarged graph of $p_i$ and three of its neighbors, $p_i$ and its projection $h_i$ on the fitted plane $H$. Note that $h_i$ is also the center of the ring formed by the virtual neighbors.}
\label{proof_graph}
\end{figure*}

\section{Curvature approximation proof}\label{ap:proof}

\begin{theorem}
Let $p_i\in \mathbb{R}^3$ be a point in point cloud. Let $H=\{x:\langle x,\vn\rangle+D=0,x\in\mathbb{R}^3\},\vn\in\mathbb{R}^3,||\vn||=1$ be the plane fitted to the neighbors of $p_i$. Let $h_i$ be the projection of $p_i$ on $H$. Assuming $p_i$'s neighbors distribute evenly and densely on a ring surrounding $h_i$, then the curvature normal at $p_i$ can be approximated by the expression $\dfrac{1}{2k^2}(h_i-p_i)$, where $k$ is the distance from $p_i$ to any of its neighbor.
\end{theorem}

\begin{proof}
Our proof will refer to Fig. \ref{proof_graph}. 

\cite{tangent} has already showed that on a 3-D mesh, given a point $p_i$ and its neighbors, the local ``carvature normal" can be calculated using
\begin{equation}\label{eq:tangent}
\dfrac{1}{4A}\sum_{j\in\mathcal{N}(p_i)}(\cot \alpha_j+\cot \beta_j)(p_j-p_i)
\end{equation}
where $A$ is the sum of the areas of the triangles having $p_i$ as common vertex and $\alpha_j,\beta_j$ are the two angles opposite to the edge $e_{ij}$ (i.e. $p_j-p_i$). This arrangement is demonstrated Fig. \ref{proof_graph}.

Since point cloud data are usually sparse and noisy, we want to utilize some mechanism to mitigate this sparsity and irregularity. Here, we first fit a local plane to $p_i$'s neighborhood, and then we define the notion of ``virtual neighbors" as a means to fill in the gaps left by the ``actual neighbors". We assume the ``virtual neighbors" distribute evenly and densely on a ring surrounding $h_i$ on the fitted plane $H$, each having the same distance $k$ to $p_i$ ($k$ is calculated using the average distance of the actual neighbors). Let $a$ be the distance from $p_i$ to each edge $e_{j,j+1}$. Let $b$ be half of the length of $e_{j,j+1}$. Thus we can calculate $A$ in Eq. \ref{eq:tangent} as $n\cdot ab$. Since we assumed the points are distributed evenly, we have $\cot \alpha=\cot \beta=\dfrac{b}{a}$. Thus we have the curvature normal to be $\dfrac{1}{4A}\sum_j(\cot \alpha_j+\cot \beta_j)(p_j-p_i)=\dfrac{1}{4nab}\cdot \dfrac{2b}{a}\sum_j(p_j-p_i)=\dfrac{1}{2na^2}\sum_j(p_j-p_i)$.

Note that the vector $p_j-p_i$ is equal to $(p_i-h_i)+(h_i-p_i)$, and it can be easily shown that $\sum_j (p_i-h_i)=\vec{0}$. Thus we can continue derive the curvature normal to be $\dfrac{1}{2na^2}\sum_j(p_j-p_i)=\dfrac{1}{2na^2}\sum_j(h_i-p_i)=\dfrac{n}{2na^2}(h_i-p_i)=\dfrac{1}{2a^2}(h_i-p_i)$. Since we assume the points are distributed densely, thus we have as $n\rightarrow \infty$, $a\rightarrow k$. Hence, the curvature normal at $p_i$ can be approximated by the expression
\begin{equation}\label{with_k}
\dfrac{1}{2k^2}(h_i-p_i)
\end{equation}
where $h_i-p_i$ is just the vector pointing from $p_i$ to the local plane $H$ as in Eq. \ref{eq:potato}. This equation makes sense in that when the distance from $p_i$ to $H$ is fixed, the further away the neighbors are, the ``flatter" the surface at $p_i$ is. 
\end{proof}

In our actual experimentation however, we found that due to the extremely irregular distribution of the point cloud data, the neighborhood distance is misleading sometimes rather than helpful. Thus, in our final algorithm, we abandon the distance information $\dfrac{1}{2k^2}$ and directly use the vector pointing from $p_i$ to plane $H$ as our approximation for the local curvature.

\section{Implementation Details for Point cloud smoothing}\label{ap:details}

\begin{figure}[htb]
\begin{subfigure}{.4\linewidth}
\centering
\includegraphics[width=0.5\linewidth]{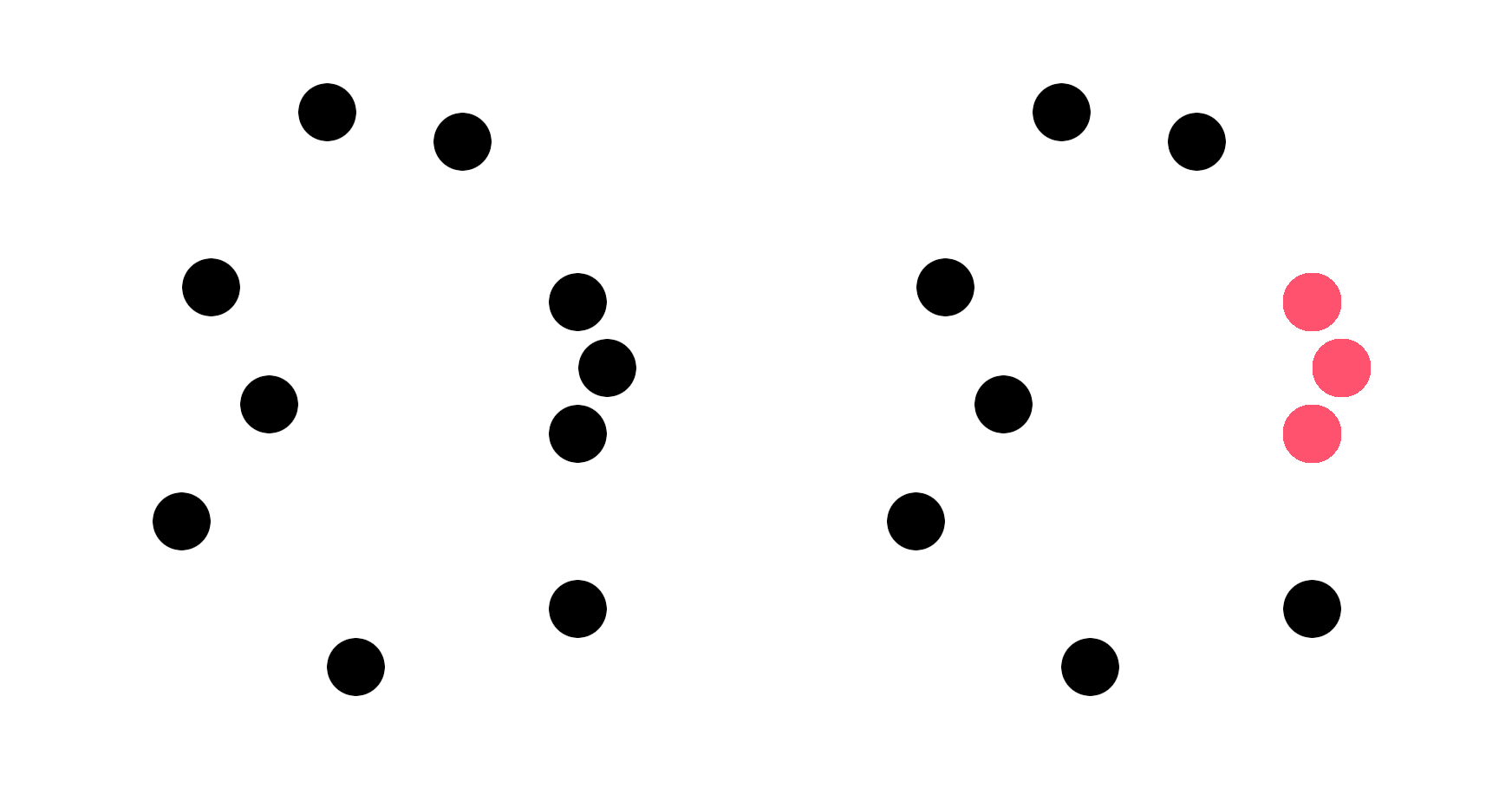}
\captionof{figure}{The ``isolated neighborhood" ($K=2$). Red dots are closed under $\mathcal{N}(\cdot)$ operation, losing contact with other points.}
\label{isolate}
\end{subfigure}\hfill
\begin{subfigure}{.55\linewidth}
\centering

\includegraphics[width=0.5\linewidth]{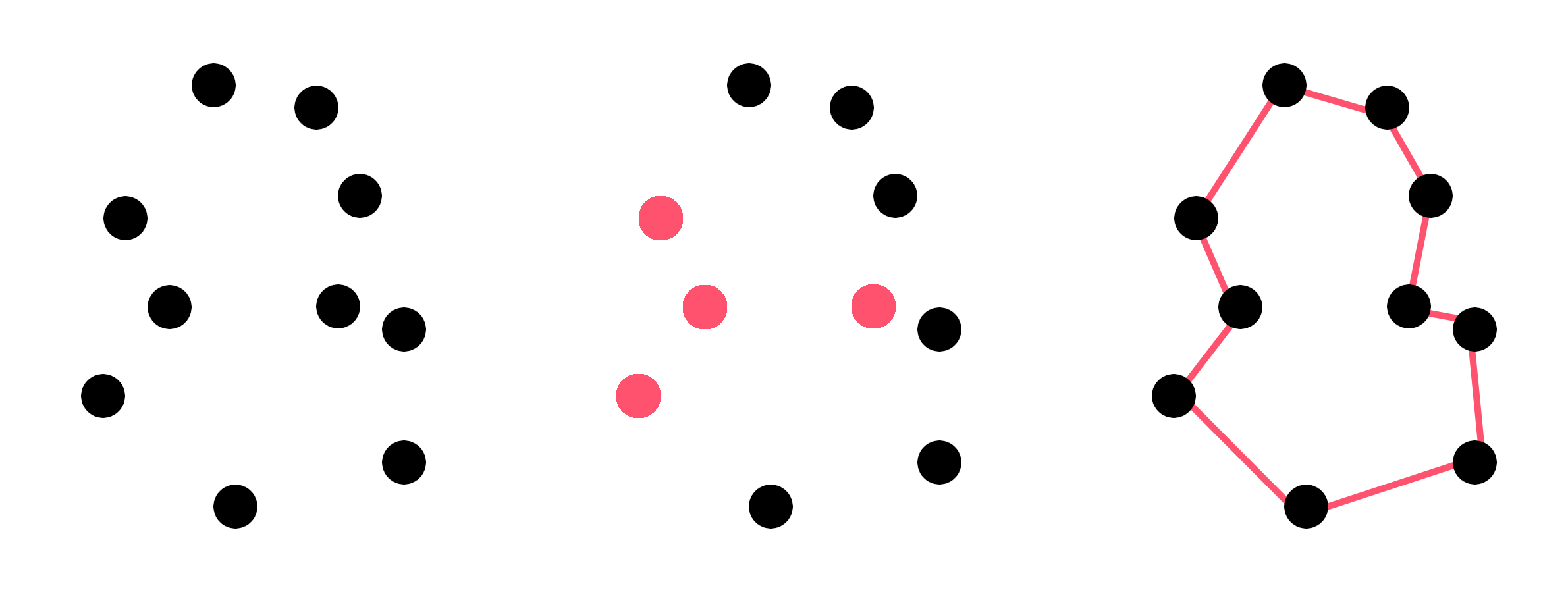}
\captionof{figure}{The ``false neighbor" ($K=3$). The rightmost red dot is a false neighbor for the red dot in the center, as indicated by the line drawing on the right.}
\label{false}
\end{subfigure}
\vspace{0.1in}
\caption{Two issues innate to point clouds (due to the missing edge information between vertices).}
\vskip -0.1in
\label{issue}
\end{figure}

An important implementation detail for the point cloud smoothing algorithm is that the size of the neighborhood we use increases as the smoothing goes further. In practice, after every 4 rounds of erosion and dilation, we expand the neighborhood size by $20$ points.
The reason for this is twofold. On one hand, there might exist isolated neighborhoods in a point cloud (i.e. a set of points that is closed under the $\mathcal{N}(\cdot)$ operation) as shown in Fig. \ref{isolate}. If the curvature information cannot be propagated to the entire point cloud, the algorithm will fail. On the other hand, a larger neighborhood speeds up the smoothing process. As mentioned in \cite{kcvs}, the time step restriction $(0<\lambda<1)$ results in the need of hundreds of updates to cause a noticeable smoothing using the original implementation in \cite{taubin2}. Note that however, we also cannot make the neighborhood size too large, especially at the beginning, due to the ``false neighbor" issue innate to the point cloud data structure (explained in Fig. \ref{false}).

\section{Smoothing Algorithm Evaluation Metrics}

\begin{figure}[htb]
\centering
\includegraphics[width=0.8\linewidth]{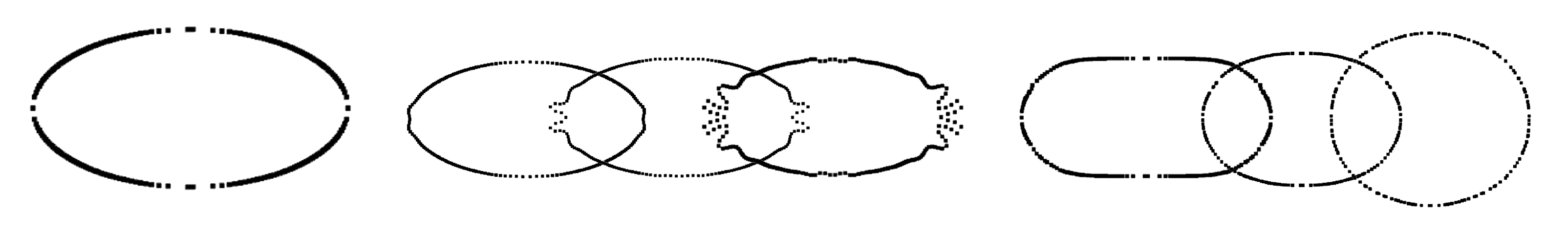}
%\vskip -0.1in
\caption{\small Left: A 2-D ellipse shape with 202 unevenly distributed points. Middle: Taubin smoothing. Right: Our smoothing. In the case of Taubin smoothing, highly concentrated areas are pushing points outward, resulting in an undesired shape (i.e., more frequent change in curvature), while our algorithm is not influenced by point density (and thus, the constant curvature is achieved as desired).}
%\vskip -0.1in
\label{ellipse}
\end{figure}

\begin{figure}[htb]
%\vskip -0.1in
\begin{minipage}{.48\linewidth}
\begin{center}
\includegraphics[width=0.95\linewidth]{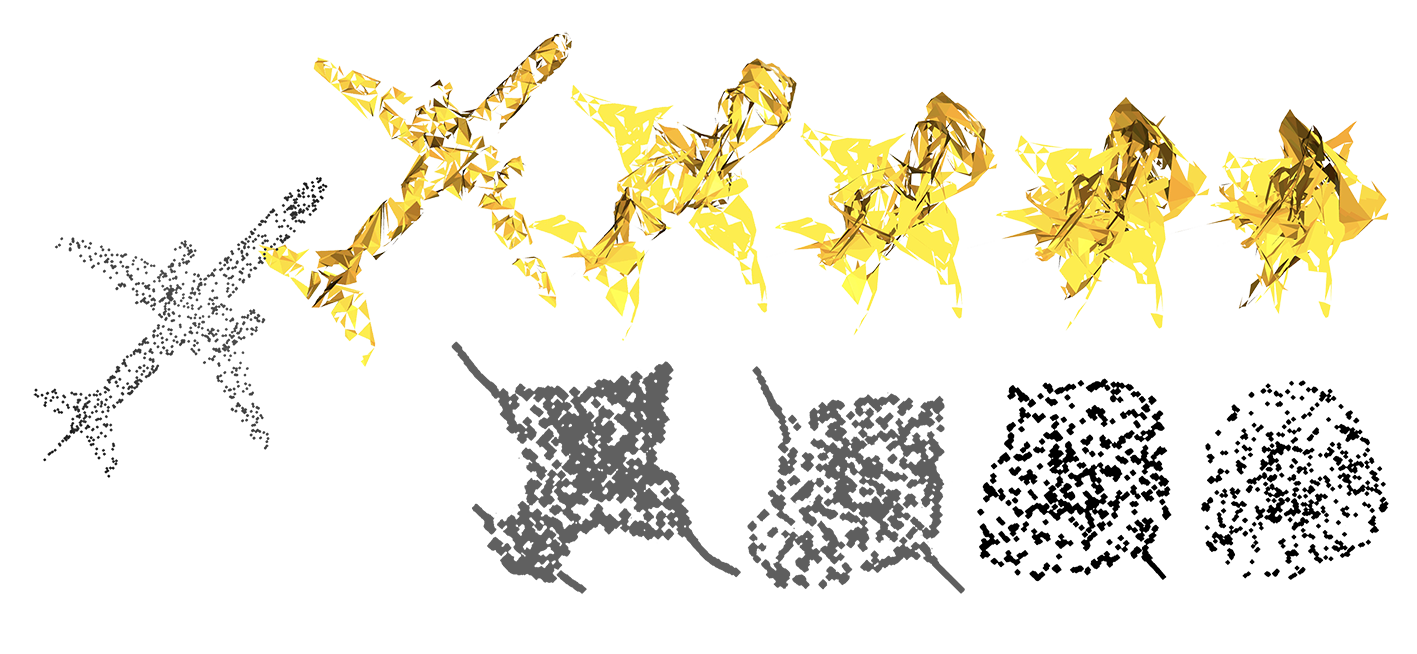}
\end{center}
%\vskip -0.2in
\captionof{figure}{Upper row: meshing a point cloud and then applying Laplacian smoothing. Lower row: our smoothing algorithm.}
\label{meshing}
\end{minipage}
\begin{minipage}{.48\linewidth}
\begin{center}
\includegraphics[width=0.95\linewidth]{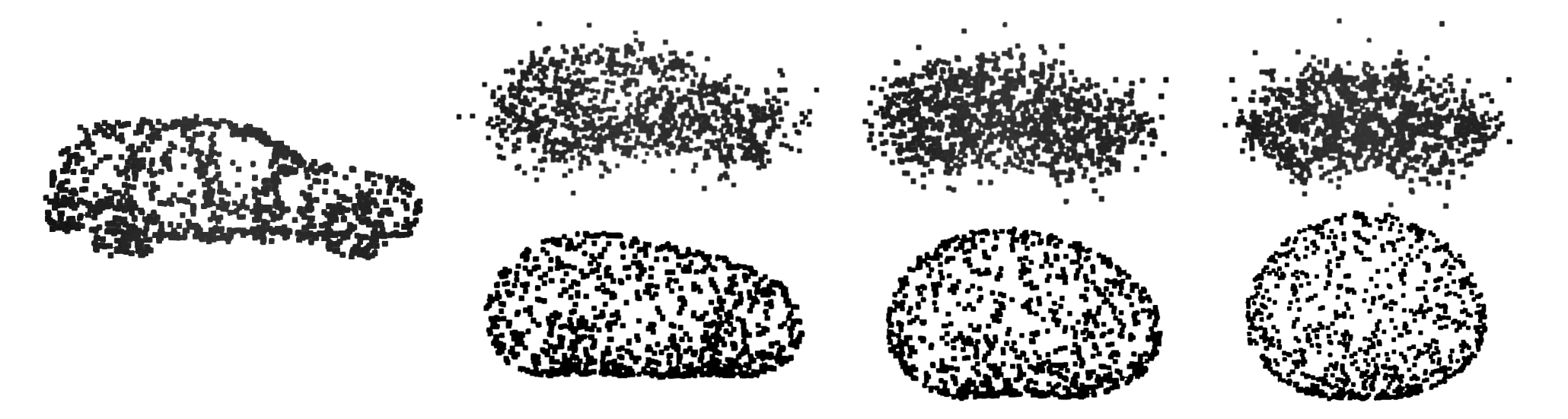}
\end{center}
%\vskip -0.2in
\caption{Upper row: mean-curvature-flow algorithm utilizing curvature from fitted quadratic surfaces to local neighborhoods. Lower row: our smoothing algorithm.}
\label{quadric}
\end{minipage}
%\vskip -0.1in
\end{figure}

In the experiment section of our paper, we use the following three metrics to compare our smoothing method against the baselines (Fig. \ref{ellipse}, \ref{meshing} and \ref{quadric} show some example comparisons in pictures):

\textbf{CSD.} Curvature standard deviation. We regard large curvature on a shape as ``features", and we want to eliminate those ``features" through the smoothing algorithms. As we remove the most distinct curvatures on the surface like edges and corners, the standard deviation of the curvatures will decrease, due to the elimination of the large outliers. In our experiment, we measure the distance from each point to its locally fitted plane ($K=60$) as an approximation of the local mean curvature.

\textbf{MR.} Min-max ratio. Assuming that the underlying manifold is closed, our smoothing should eventually morph the point cloud into a sphere. Hence, we propose to evaluate the ratio between the length on the short side and the long side of the point cloud. This is computed by first applying principal component analysis (PCA) to the point cloud and finding the top two principal components, say $\vec{u}$ and $\vec{v}$. Then we compute the ratio between ranges of the values on these two principal directions.
The closer this ratio is to 1, the better.

\textbf{DDS.} Density distribution similarity. We want the morphing process to be smooth in that the density distribution of each point cloud to \textit{remain the same} throughout the morphing process. We conduct the kernel density estimation at each point (using a Gaussian kernel with $\sigma=0.1$) to obtain the density distribution of the entire point cloud, and then compare the similarity between the distributions of two consecutive blurred levels using the Kolmogorov-Smirnov test (p-values are recorded as results). 

\newpage
\section{Curve figures}\label{ap:curves}

%%%%%%%%%pointconv
\begin{figure*}[h]

\begin{subfigure}{.24\linewidth}
\includegraphics[width=0.95\linewidth]{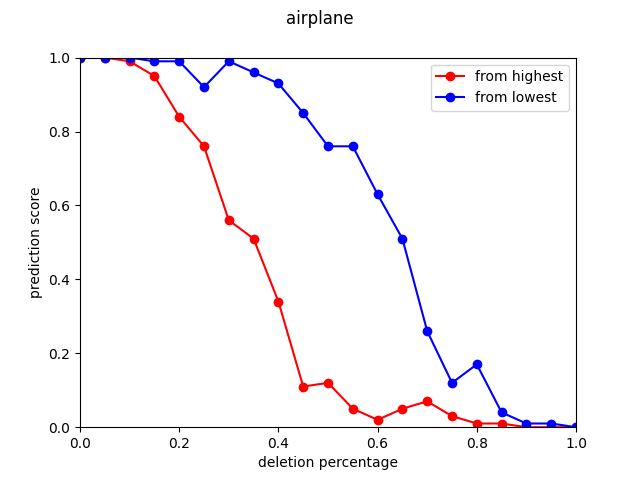}
\captionof{figure}{Airplane.}
\end{subfigure}\hfill
\begin{subfigure}{.20\linewidth}
\includegraphics[width=0.95\linewidth]{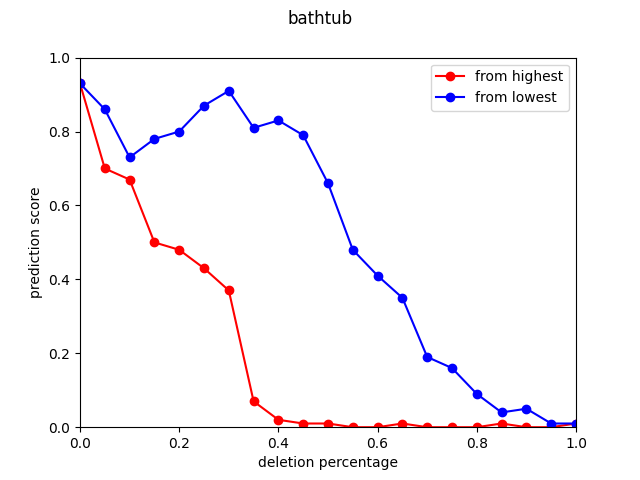}
\captionof{figure}{Bathtub.}
\end{subfigure}\hfill
\begin{subfigure}{.20\linewidth}
\includegraphics[width=0.95\linewidth]{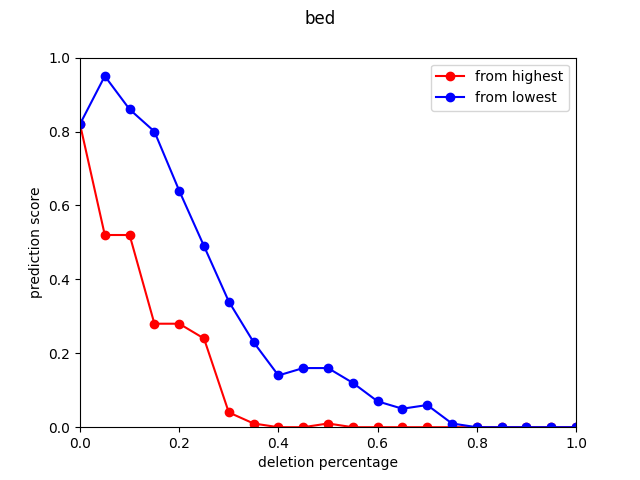}
\captionof{figure}{Bed.}
\end{subfigure}\hfill
\begin{subfigure}{.20\linewidth}
\includegraphics[width=0.95\linewidth]{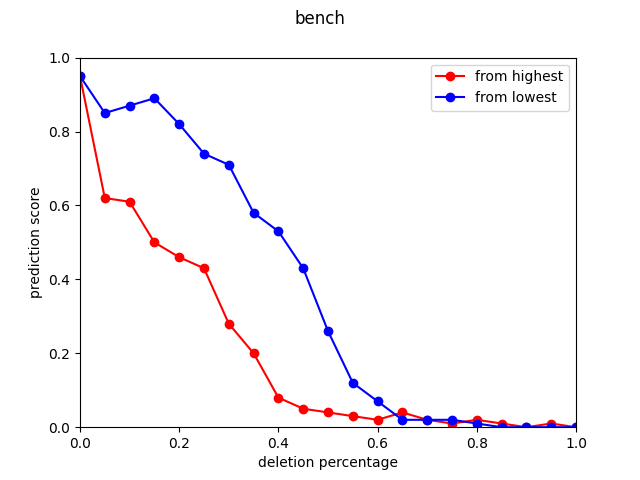}
\captionof{figure}{Bench.}
\end{subfigure}\hfill

\begin{subfigure}{.20\linewidth}
\includegraphics[width=0.95\linewidth]{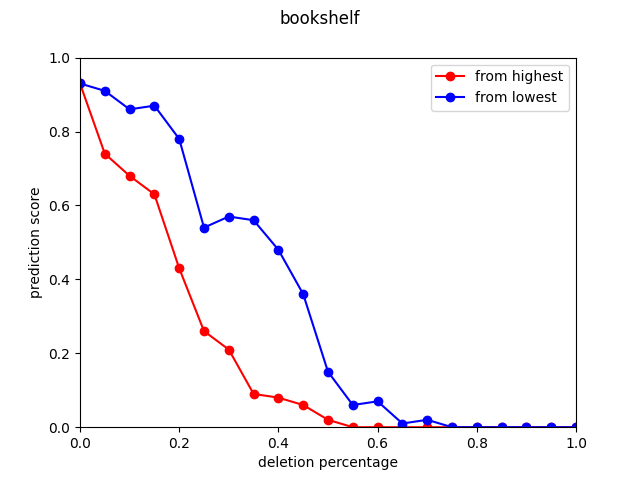}
\captionof{figure}{Bookshelf.}
\end{subfigure}\hfill
\begin{subfigure}{.20\linewidth}
\includegraphics[width=0.95\linewidth]{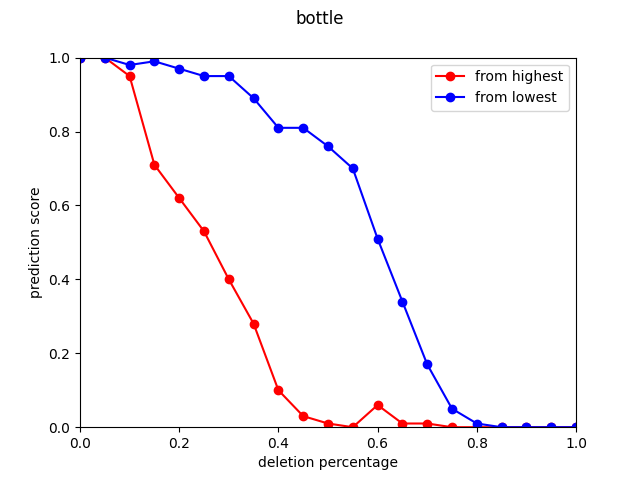}
\captionof{figure}{Bottle.}
\end{subfigure}\hfill
\begin{subfigure}{.20\linewidth}
\includegraphics[width=0.95\linewidth]{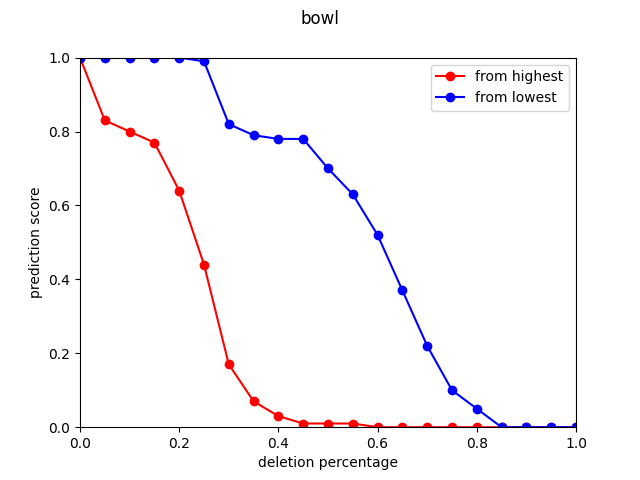}
\captionof{figure}{Bowl.}
\end{subfigure}\hfill
\begin{subfigure}{.20\linewidth}
\includegraphics[width=0.95\linewidth]{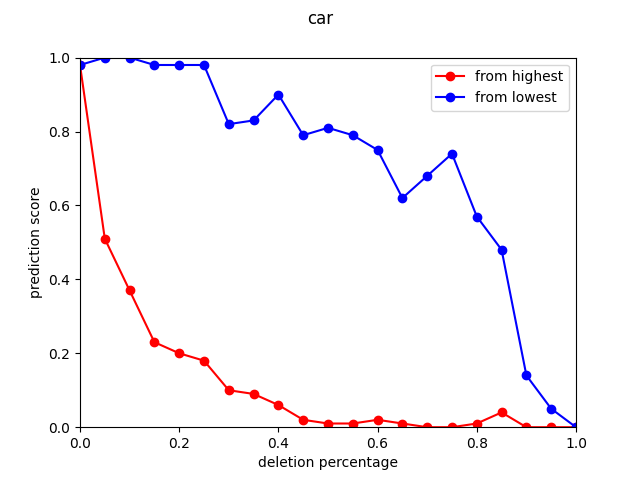}
\captionof{figure}{Car.}
\end{subfigure}\hfill

\begin{subfigure}{.20\linewidth}
\includegraphics[width=0.95\linewidth]{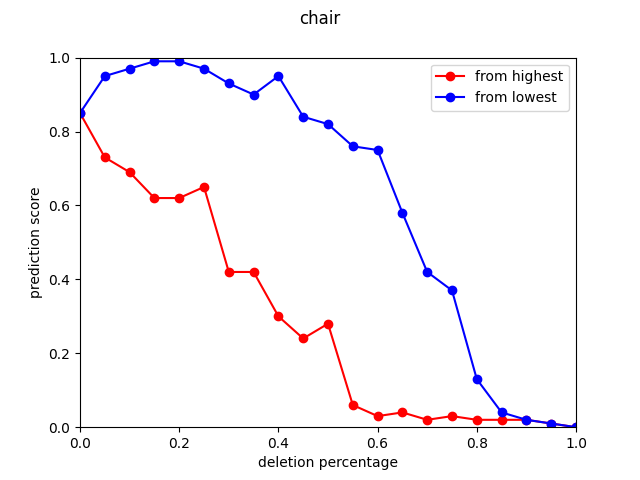}
\captionof{figure}{Chair.}
\end{subfigure}\hfill
\begin{subfigure}{.20\linewidth}
\includegraphics[width=0.95\linewidth]{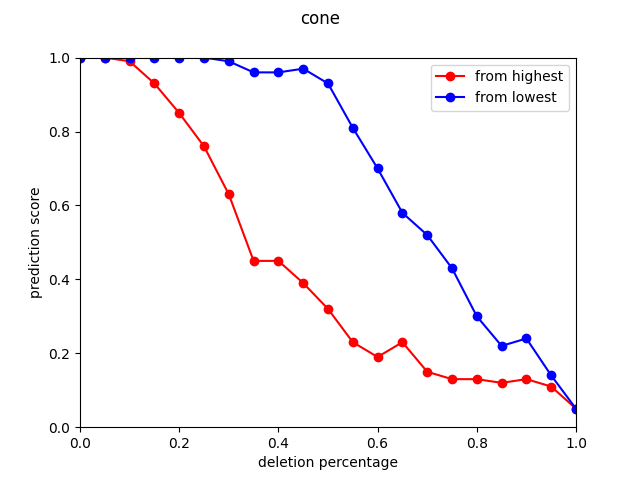}
\captionof{figure}{Cone.}
\end{subfigure}\hfill
\begin{subfigure}{.20\linewidth}
\includegraphics[width=0.95\linewidth]{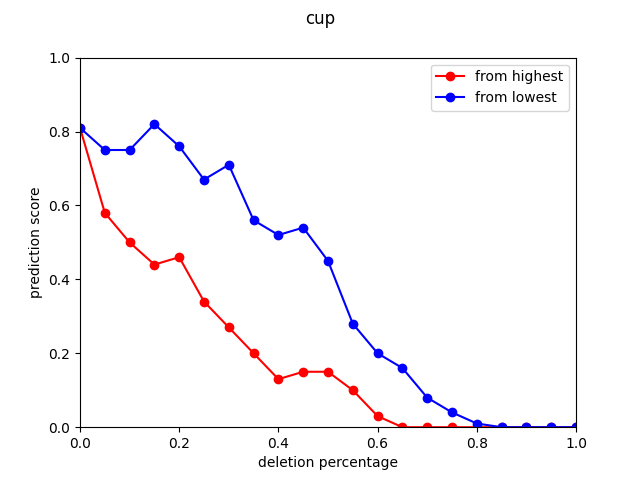}
\captionof{figure}{Cup.}
\end{subfigure}\hfill
\begin{subfigure}{.20\linewidth}
\includegraphics[width=0.95\linewidth]{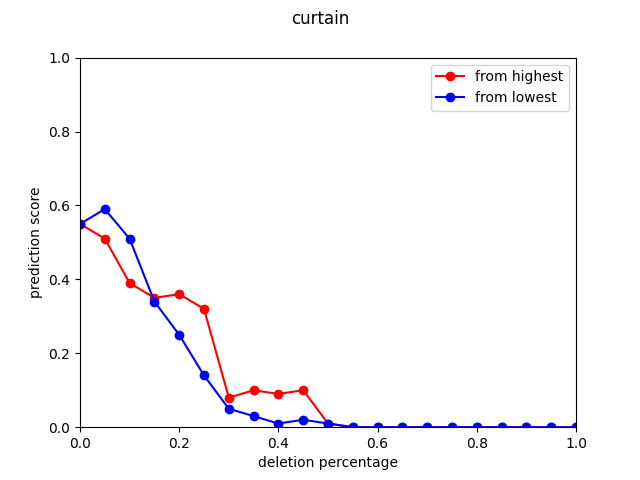}
\captionof{figure}{Curtain.}
\end{subfigure}\hfill

\begin{subfigure}{.20\linewidth}
\includegraphics[width=0.95\linewidth]{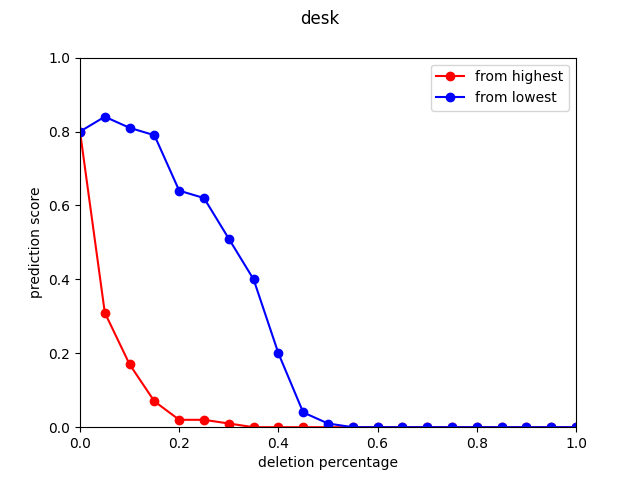}
\captionof{figure}{Desk.}
\end{subfigure}\hfill
\begin{subfigure}{.20\linewidth}
\includegraphics[width=0.95\linewidth]{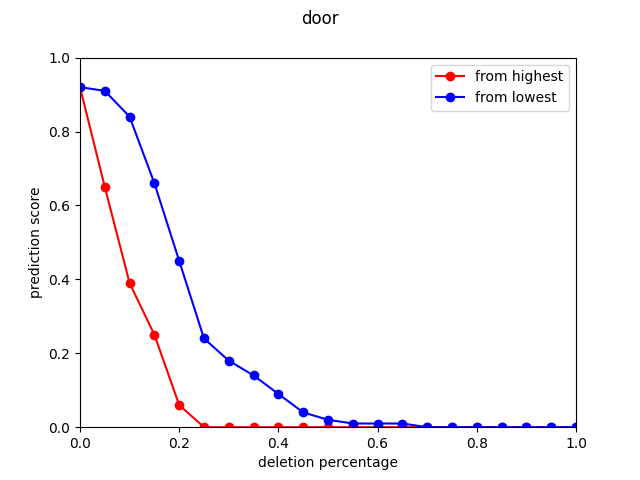}
\captionof{figure}{Door.}
\end{subfigure}\hfill
\begin{subfigure}{.20\linewidth}
\includegraphics[width=0.95\linewidth]{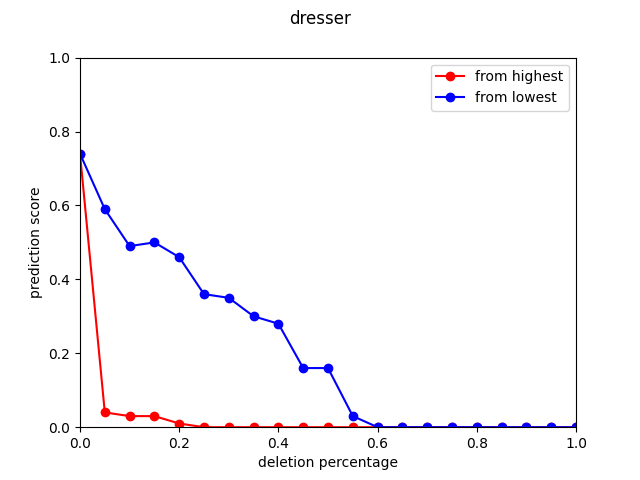}
\captionof{figure}{Dresser.}
\end{subfigure}\hfill
\begin{subfigure}{.20\linewidth}
\includegraphics[width=0.95\linewidth]{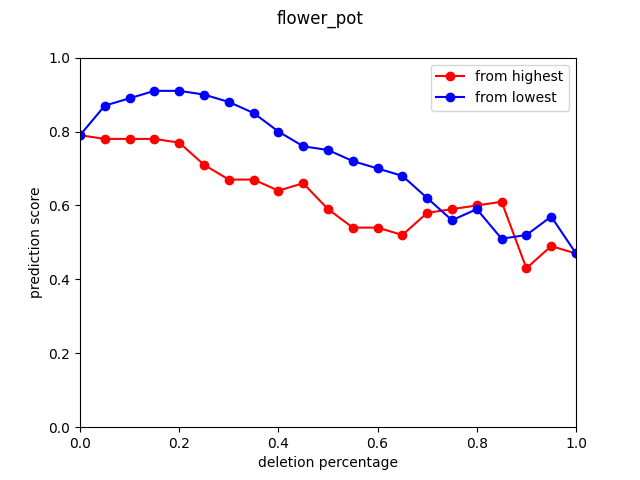}
\captionof{figure}{Flower pot.}
\end{subfigure}\hfill

\begin{subfigure}{.20\linewidth}
\includegraphics[width=0.95\linewidth]{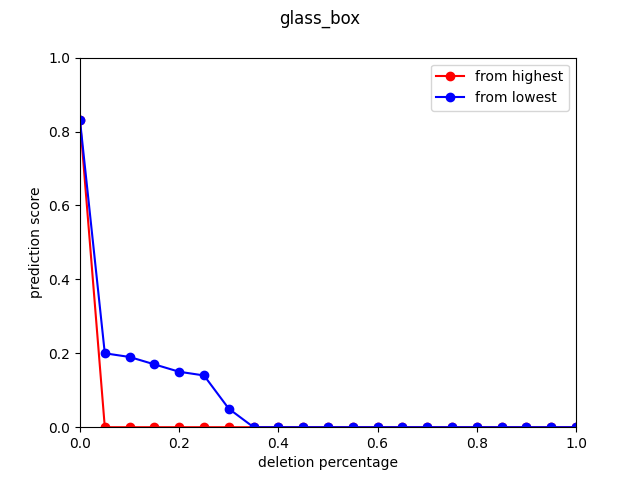}
\captionof{figure}{Glass box.}
\end{subfigure}\hfill
\begin{subfigure}{.20\linewidth}
\includegraphics[width=0.95\linewidth]{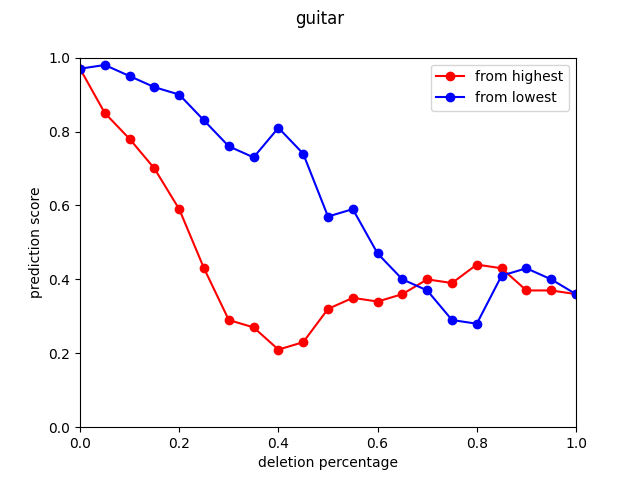}
\captionof{figure}{Guitar.}
\end{subfigure}\hfill
\begin{subfigure}{.20\linewidth}
\includegraphics[width=0.95\linewidth]{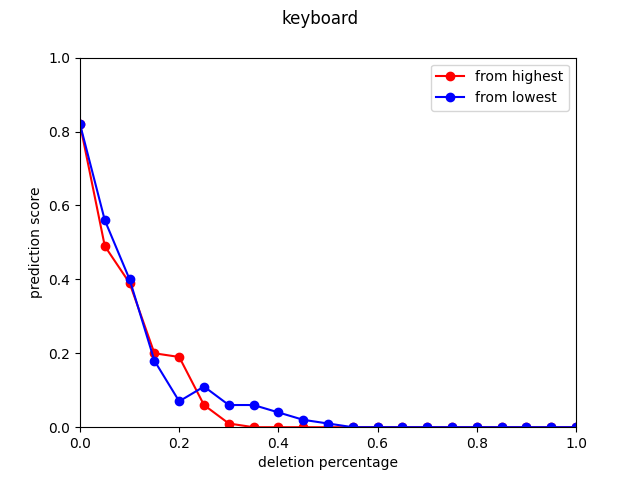}
\captionof{figure}{Keyboard.}
\end{subfigure}\hfill
\begin{subfigure}{.20\linewidth}
\includegraphics[width=0.95\linewidth]{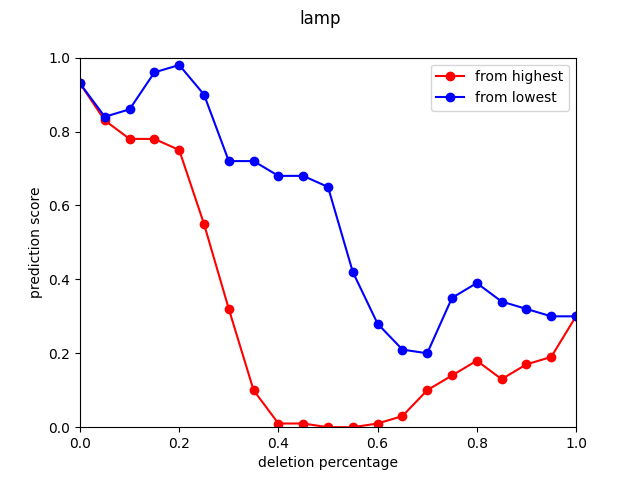}
\captionof{figure}{Lamp.}
\end{subfigure}\hfill

\vspace{0.1in}
\caption{\textit{Deletion} and \textit{insertion} curves for all 40 classes in ModelNet40 for PointConv. Horizontal axis is the deletion percentage (top 5\%, 10\%, etc.), and vertical axis is the predicted class score. The red line is the \textit{deletion} curve which blurs points from highest mask values, and the blue line is the \textit{insertion} curve (if read from right to left) which blurs points from lowest mask values.}
\end{figure*}

\begin{figure*}[h]\ContinuedFloat

\begin{subfigure}{.20\linewidth}
\includegraphics[width=0.95\linewidth]{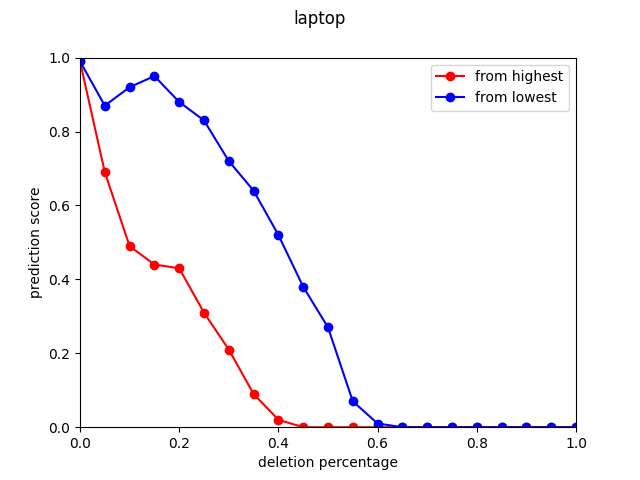}
\captionof{figure}{Laptop.}
\end{subfigure}\hfill
\begin{subfigure}{.20\linewidth}
\includegraphics[width=0.95\linewidth]{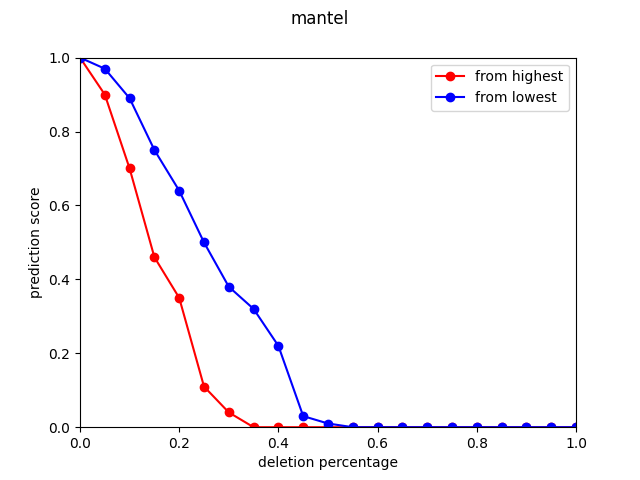}
\captionof{figure}{Mantel.}
\end{subfigure}\hfill
\begin{subfigure}{.20\linewidth}
\includegraphics[width=0.95\linewidth]{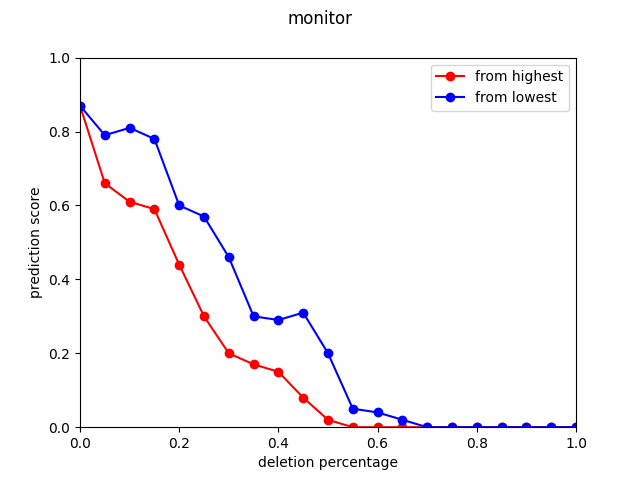}
\captionof{figure}{Monitor.}
\end{subfigure}\hfill
\begin{subfigure}{.20\linewidth}
\includegraphics[width=0.95\linewidth]{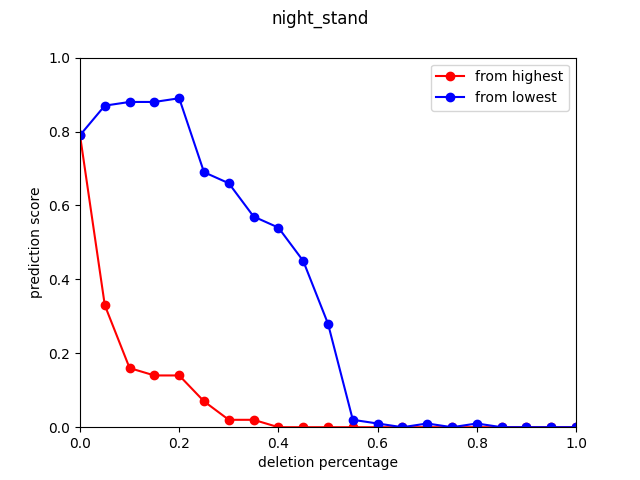}
\captionof{figure}{Night stand.}
\end{subfigure}\hfill

\begin{subfigure}{.20\linewidth}
\includegraphics[width=0.95\linewidth]{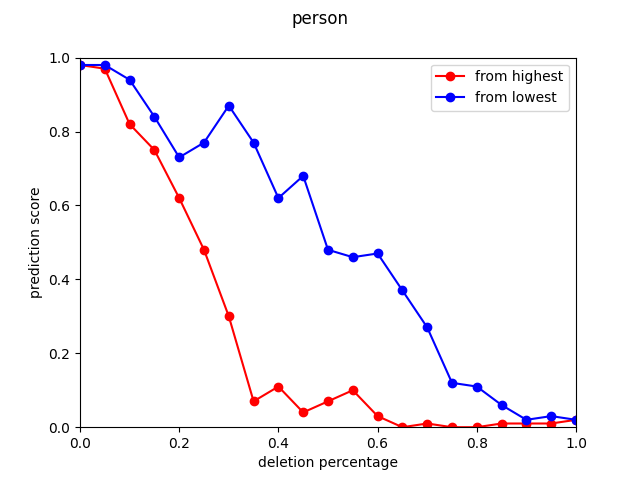}
\captionof{figure}{Person.}
\end{subfigure}\hfill
\begin{subfigure}{.20\linewidth}
\includegraphics[width=0.95\linewidth]{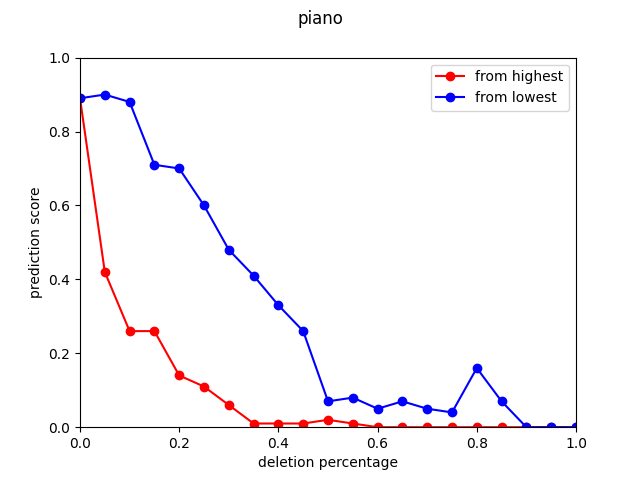}
\captionof{figure}{Piano.}
\end{subfigure}\hfill
\begin{subfigure}{.20\linewidth}
\includegraphics[width=0.95\linewidth]{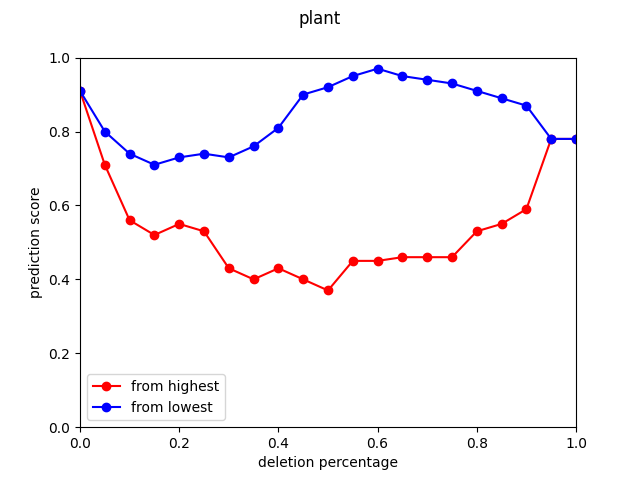}
\captionof{figure}{Plant.}
\end{subfigure}\hfill
\begin{subfigure}{.20\linewidth}
\includegraphics[width=0.95\linewidth]{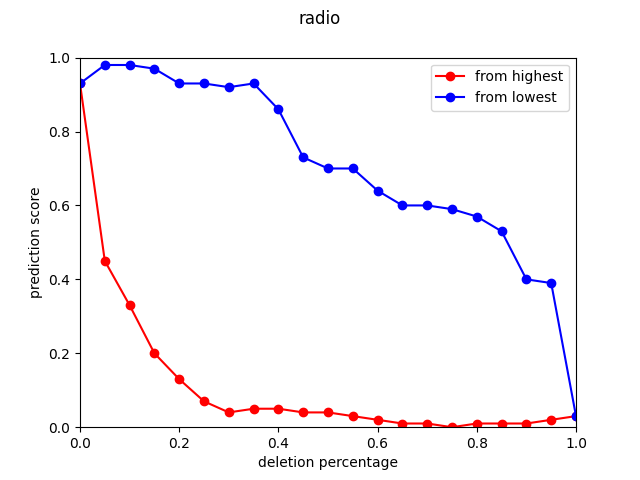}
\captionof{figure}{Radio.}
\end{subfigure}\hfill

\begin{subfigure}{.20\linewidth}
\includegraphics[width=0.95\linewidth]{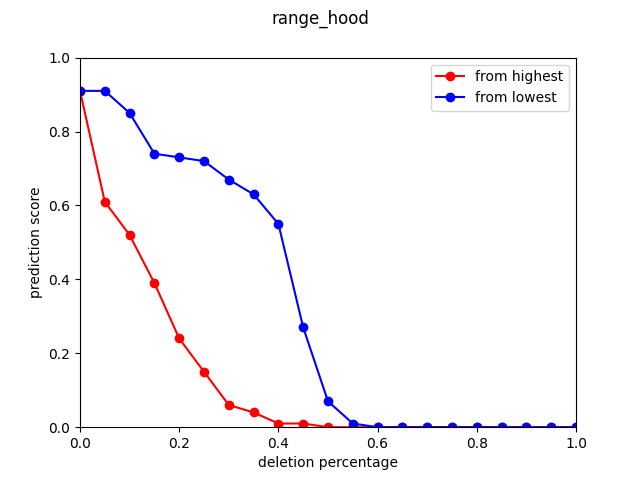}
\captionof{figure}{Range hood.}
\end{subfigure}\hfill
\begin{subfigure}{.20\linewidth}
\includegraphics[width=0.95\linewidth]{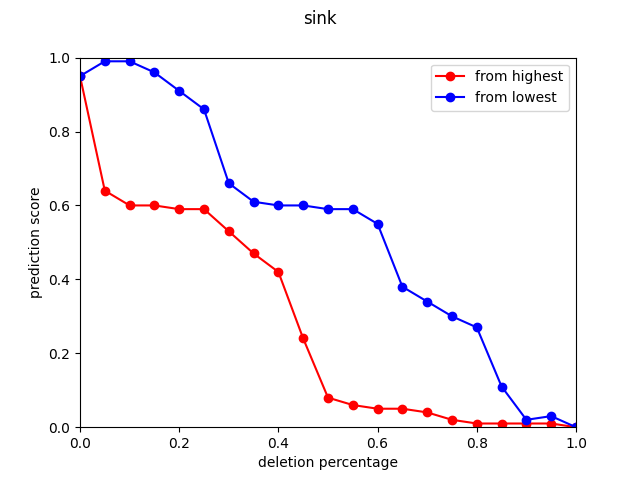}
\captionof{figure}{Sink.}
\end{subfigure}\hfill
\begin{subfigure}{.20\linewidth}
\includegraphics[width=0.95\linewidth]{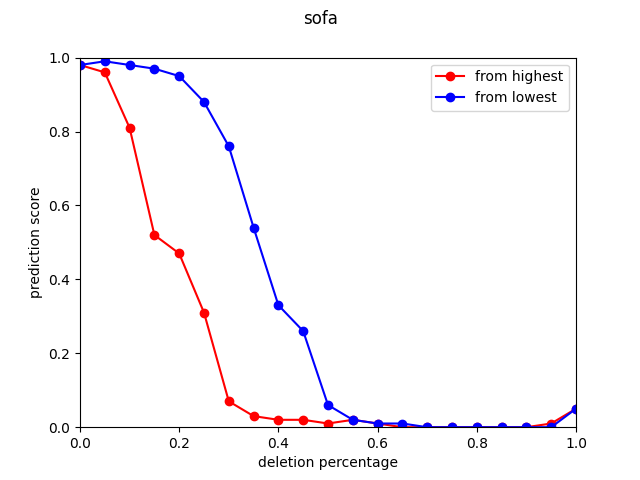}
\captionof{figure}{Sofa.}
\end{subfigure}\hfill
\begin{subfigure}{.20\linewidth}
\includegraphics[width=0.95\linewidth]{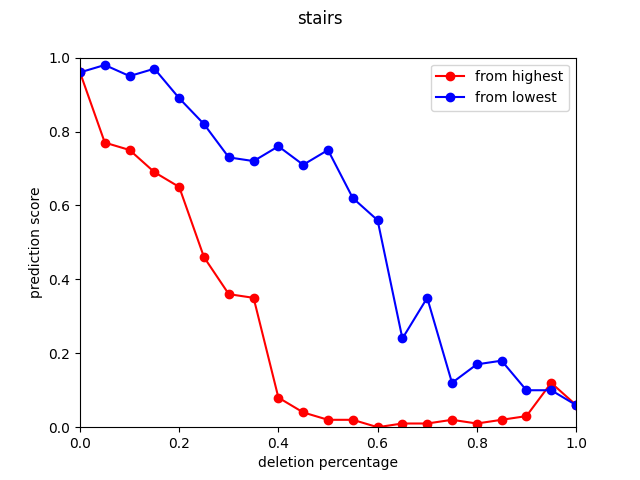}
\captionof{figure}{Stairs.}
\end{subfigure}\hfill

\begin{subfigure}{.20\linewidth}
\includegraphics[width=0.95\linewidth]{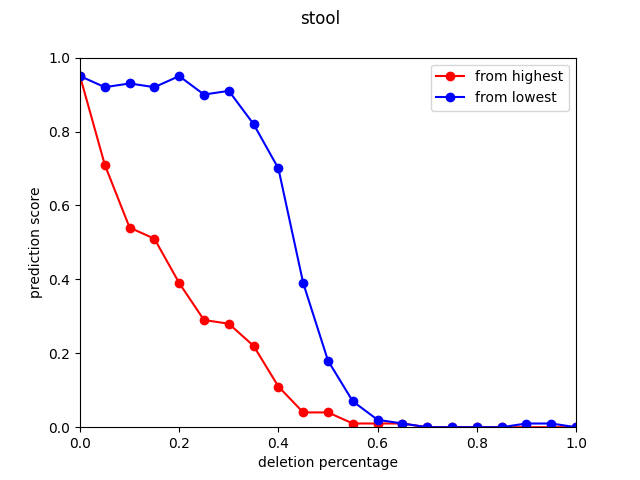}
\captionof{figure}{Stool.}
\end{subfigure}\hfill
\begin{subfigure}{.20\linewidth}
\includegraphics[width=0.95\linewidth]{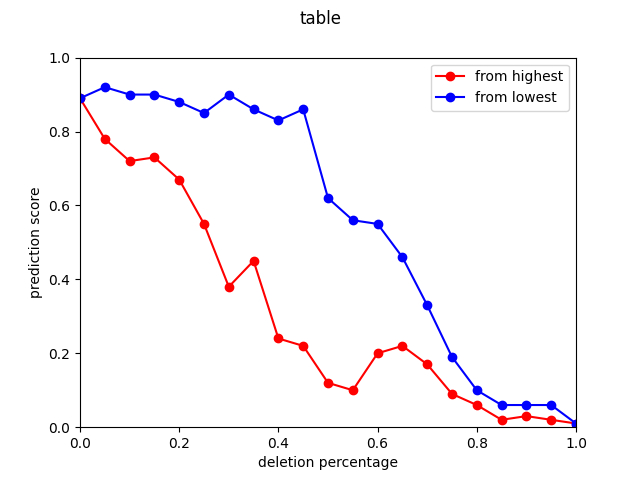}
\captionof{figure}{Table.}
\end{subfigure}\hfill
\begin{subfigure}{.20\linewidth}
\includegraphics[width=0.95\linewidth]{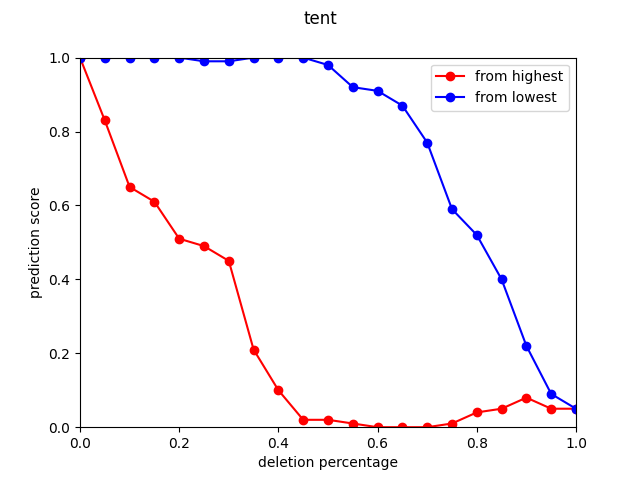}
\captionof{figure}{Tent.}
\end{subfigure}\hfill
\begin{subfigure}{.20\linewidth}
\includegraphics[width=0.95\linewidth]{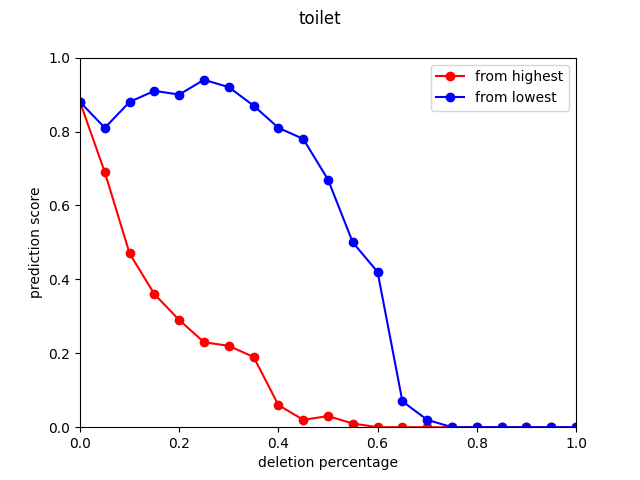}
\captionof{figure}{Toilet.}
\end{subfigure}\hfill

\begin{subfigure}{.20\linewidth}
\includegraphics[width=0.95\linewidth]{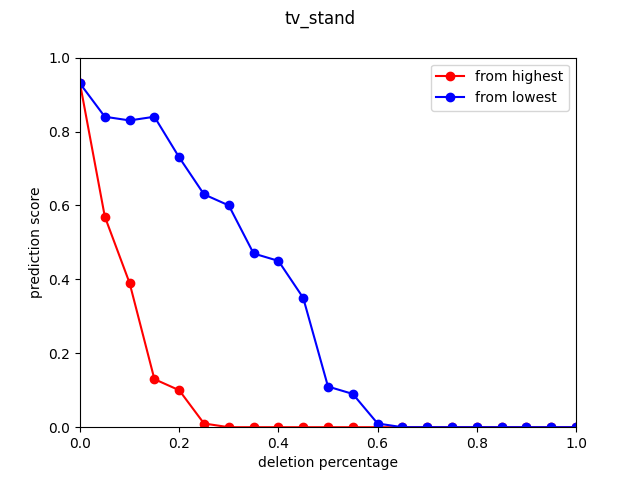}
\captionof{figure}{TV stand.}
\end{subfigure}\hfill
\begin{subfigure}{.20\linewidth}
\includegraphics[width=0.95\linewidth]{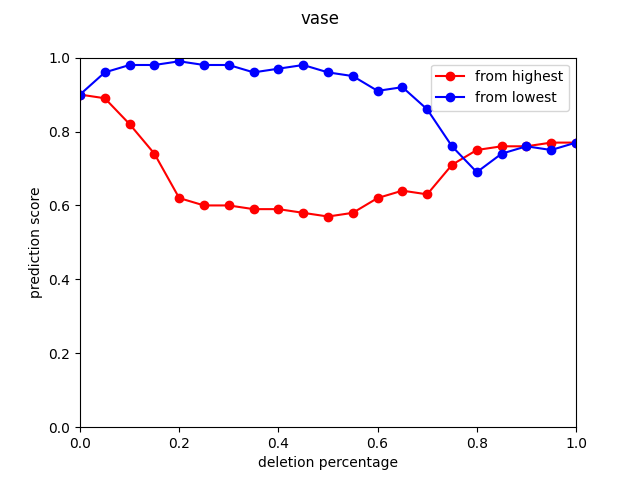}
\captionof{figure}{Vases.}
\end{subfigure}\hfill
\begin{subfigure}{.20\linewidth}
\includegraphics[width=0.95\linewidth]{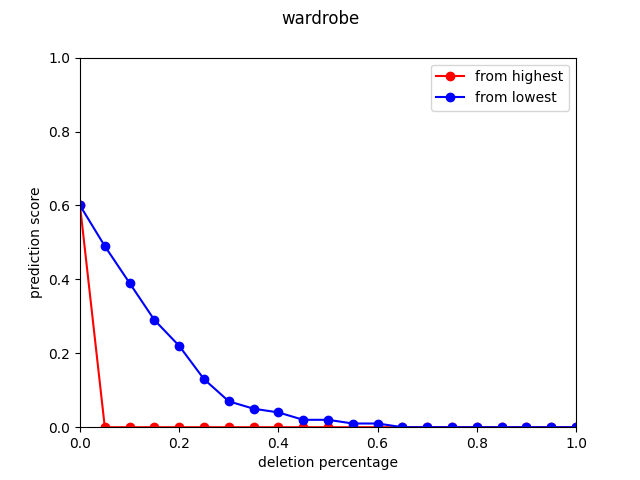}
\captionof{figure}{Wardrobe.}
\end{subfigure}\hfill
\begin{subfigure}{.20\linewidth}
\includegraphics[width=0.95\linewidth]{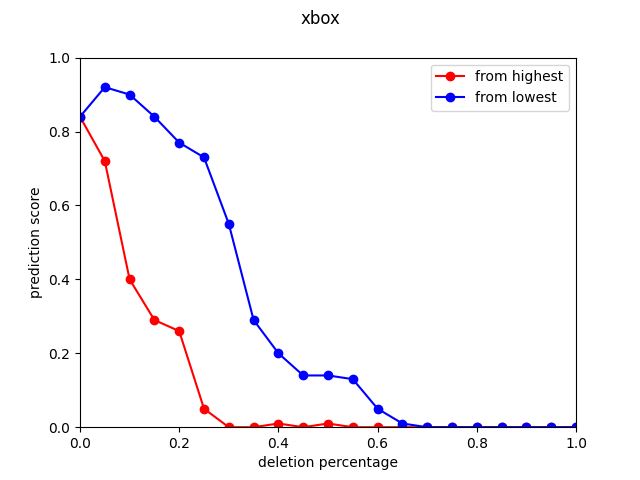}
\captionof{figure}{Xbox.}
\end{subfigure}\hfill

\vspace{0.1in}
\caption{\textit{Deletion} and \textit{insertion} curves for all 40 classes in ModelNet40 for PointConv. Horizontal axis is the deletion percentage (top 5\%, 10\%, etc.), and vertical axis is the predicted class score. The red line is the \textit{deletion} curve which blurs points from highest mask values, and the blue line is the \textit{insertion} curve (if read from right to left) which blurs points from lowest mask values. (cont.)}
\label{pc_curves}
\end{figure*}

\clearpage

\bibliography{egbib}

\end{document}